# Developing a Knowledge Graph Framework for Pharmacokinetic Natural Product-Drug Interactions

Sanya B. Taneja*[1], Tiffany J. Callahan[2], Mary F. Paine[3], Sandra L. Kane-Gill[4], Halil Kilicoglu[5], Marcin P. Joachimiak[6], Richard D. Boyce[7]

[1]Intelligent Systems Program, University of Pittsburgh, Pittsburgh, PA 15206, USA
[2]Department of Biomedical Informatics, Columbia University, New York, NY 10032, USA
[3]Department of Pharmaceutical Sciences, College of Pharmacy and Pharmaceutical Sciences, Washington State University, Spokane, WA 99202, USA
[4]School of Pharmacy, University of Pittsburgh, Pittsburgh, PA 15261, USA
[5]School of Information Sciences, University of Illinois at Urbana-Champaign, Champaign, IL 61820, USA
[6]Environmental Genomics and Systems Biology Division, Lawrence Berkeley National Laboratory, Berkeley, CA 94720, USA
[7]Department of Biomedical Informatics, University of Pittsburgh, Pittsburgh, PA 15206, USA

*Corresponding Author

**Keywords**
Biomedical ontology, knowledge graph, knowledge representation, literature-based discovery, natural products, pharmacokinetics

**Abstract**

*Background*: Pharmacokinetic natural product-drug interactions (NPDIs) occur when botanical or other natural products are co-consumed with pharmaceutical drugs. With the growing use of natural products, the risk for potential NPDIs and consequent adverse events has increased. Understanding mechanisms of NPDIs is key to preventing or minimizing adverse events. Although biomedical knowledge graphs (KGs) have been widely used for drug-drug interaction applications, computational investigation of NPDIs is novel. We constructed a KG framework, NP-KG, as a first step toward computational discovery of plausible mechanistic explanations for pharmacokinetic NPDIs that can be used to guide scientific research.

*Methods*: We developed a large-scale, heterogeneous KG with biomedical ontologies, linked data, and full texts of the scientific literature. To construct the KG, biomedical ontologies and drug databases were integrated with the Phenotype Knowledge Translator framework. The semantic relation extraction systems, SemRep and Integrated Network and Dynamic Reasoning Assembler, were used to extract semantic predications (subject-relation-object triples) from full texts of the scientific literature related to the exemplar natural products green tea and kratom. A literature-based graph constructed from the predications was integrated into the ontology-grounded KG to create NP-KG. NP-KG was evaluated with case studies of pharmacokinetic green tea- and kratom-drug interactions through KG path searches and meta-path discovery to determine congruent and contradictory information in NP-KG compared to ground truth data. We also conducted an error analysis to identify knowledge gaps and incorrect predications in the KG.

*Results*: The fully integrated NP-KG consisted of 745,512 nodes and 7,249,576 edges. Evaluation of NP-KG resulted in congruent (38.98% for green tea, 50% for kratom), contradictory (15.25% for green tea, 21.43% for kratom), and both congruent and contradictory (15.25% for green tea, 21.43% for kratom) information compared to ground truth data. Potential pharmacokinetic mechanisms for several purported NPDIs, including the green tea-raloxifene, green tea-nadolol, kratom-midazolam, kratom-quetiapine, and kratom-venlafaxine interactions were congruent with the published literature.

*Conclusion*: NP-KG is the first KG to integrate biomedical ontologies with full texts of the scientific literature focused on natural products. We demonstrate the application of NP-KG to identify known pharmacokinetic interactions between natural products involving enzymes, transporters, and pharmaceutical drugs. Future work will incorporate context, contradiction analysis, and embedding-based methods to enrich NP-KG. We envision that NP-KG will facilitate improved human-machine collaboration to guide researchers in future studies of pharmacokinetic NPDIs. The NP-KG framework is publicly available at https://doi.org/10.5281/zenodo.6814507. The code for relation extraction, KG construction, and hypothesis generation is available at https://github.com/sanyabt/np-kg.

1. **Introduction**

Complementary health approaches involving the use of botanical and other natural products have increased substantively since passage of the Dietary Supplement Health and Education Act in 1994. Up to 18% of adults reported regular use of natural products in the United States (US) [1], and annual sales of herbal dietary supplements in the US increased by a record 17.3% in 2020 [2]. Although not intended to replace pharmaceutical drugs, co-consuming these products with drugs is common [3,4]. Older adults are the largest consumers of both pharmaceutical drugs and natural products, with up to 88% in the US reporting concomitant use [5]. However, such concomitant use may result in pharmacokinetic natural product-drug interactions (NPDIs) and potentially, unexpected drug responses [6]. Although substantial research has been devoted to understanding mechanisms and clinical effects of drug-drug interactions, corresponding information about NPDIs remains lacking. With the safety concerns related to the growing use of natural products, understanding the mechanisms underlying their interactions with other xenobiotics is imperative to prevent or minimize potential adverse events.

Pharmacokinetic NPDIs occur when a natural product alters the absorption, distribution, metabolism, and/or excretion of a co-consumed drug, potentially resulting in reduced treatment efficacy or adverse events [7]. For example, the popular botanical product green tea (*Camellia sinensis*), which is available as a beverage and dietary supplement, precipitated an interaction with the beta-blocker and anti-hypertensive agent nadolol, possibly by inhibiting intestinal human organic anion transporting polypeptide(s) (OATPs), uptake transporters that facilitates absorption of the substrate from the gut lumen into the systemic circulation [8]. This pharmacokinetic interaction in turn led to a decrease in the blood pressure lowering effect of nadolol.

Understanding biochemical mechanisms underlying clinically significant pharmacokinetic NPDIs can help prevent or minimize adverse events [4]. Similar to drug-drug interactions, computational approaches to elucidate potential mechanisms can help answer the 'why' question; that is, why the NPDI occurs and through what mechanism(s) [9]. Prior work in computational discovery of potential NPDIs has focused on mining scientific abstracts to classify NPDIs involving dietary supplements [10] and developing a knowledge graph (KG) to identify plausible drug-supplement interactions [11]. Using scientific abstracts to classify supplement-drug interactions has shown promise; however, the model developed by Wang et. al. [10] was trained using a drug-drug interaction dataset due to lack of labeled NPDI datasets. Schutte et. al. [11] extracted knowledge from scientific abstracts to develop a literature-based KG and evaluated the graph with discovery patterns followed by a clinical review. Computational prediction has also utilized homogeneous graph structure constructed from existing resources for chemicals and foods [12]. Dietary supplements sold in the US represent a broad group of products, including botanicals, which are regulated differently than pharmaceutical drugs [13]. The US Food and Drug Administration (FDA) can prohibit the sale of these products only if they are found to be unsafe; they do not approve them prior to marketing.

Following a similar workflow used during drug development to identify pharmacokinetic drug-drug interactions [14], potential pharmacokinetic NPDIs can be evaluated by first conducting *in vitro* studies to determine if a natural product extract (e.g., some quantity of green tea) phytoconstituent (e.g., catechin) inhibits or induces the function of a metabolic enzyme or transport protein, which may or may not have unforeseen negative consequences. Results are then used to quantitatively predict the clinical impact of the interaction using *in vitro* to *in vivo* extrapolation (IVIVE). Potential clinically relevant pharmacokinetic NPDIs suggested by IVIVE are then evaluated *via* dynamic (e.g., physiologically based pharmacokinetic) modeling and simulation and/or human clinical studies [4].

Although effective, *in vitro* experiments followed by IVIVE often predict NPDIs that are not confirmed by more time-consuming and costly clinical studies. For example, a mechanistic static model predicted green tea to increase the systemic exposure to the selective estrogen receptor modulator raloxifene by inhibiting intestinal uridine 5'-diphospho-glucuronosyltransferase (UDP-glucuronosyltransferase, UGT), but the opposite was observed when tested in healthy volunteers [15,16]. Although green tea is promoted for weight loss, mental alertness, relieving headaches and digestive symptoms, and cardio protection, the popularity and widespread availability of green tea products, as well as the likelihood of co-consumption with pharmaceutical drugs increases the risk of potential clinically significant green tea-drug interactions [15].

Developing novel methods that help scientists make accurate and timely NPDI predictions is crucial to reduce the number of patients who experience adverse events from NPDIs. We hypothesized that a KG built using rigorous biomedical ontologies and enriched with domain-specific information from the scientific literature will generate plausible mechanistic explanations for pharmacokinetic NPDIs that can be used to guide robust NPDI research. In this study, we designed and built a novel KG framework, termed NP-KG, as a first step towards testing this hypothesis. We evaluated whether NP-KG accurately represented known pharmacokinetic mechanisms for NPDIs involving two exemplar natural products: green tea and kratom. Green tea was selected due to its worldwide use and the published results related to NPDIs discussed above. Kratom (*Mitragyna speciosa*) is an emerging botanical product commonly used to self-treat pain, anxiety, and opioid withdrawal symptoms. Kratom leaves contain numerous alkaloids that produce stimulant effects at low doses and opioid-like effects at higher doses. However, safety concerns have been raised by various federal agencies related to kratom toxicity and potential NPDIs [17–20]. We further demonstrated the potential of NP-KG to generate hypotheses for mechanisms underlying pharmacokinetic NPDIs related to kratom reported in published case reports.

## 2. Materials and Methods

A biomedical KG combines concepts from different domains to represent entities such as molecules, genes, diseases, and chemical substances as nodes in the graph. Edges in the KG represent relations between the entities. We describe the construction of the ontology-grounded KG, the process of semantic relation extraction from full texts of the literature, and integration into the NP-KG framework (Figure 1). Semantic relation extraction produces predications (subject-relation-object triples) from texts, where the subjects and objects can form the nodes in the KG, and relations form edges between them. Lastly, we discuss the strategies used to evaluate NP-KG.

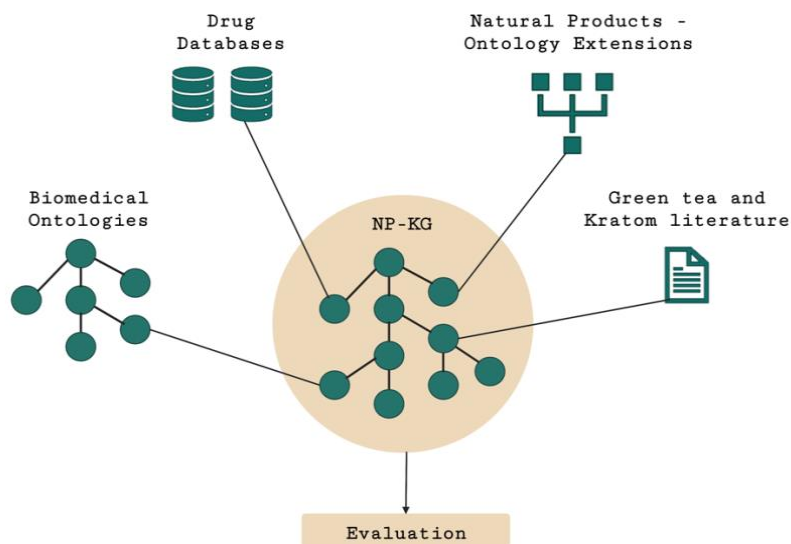

**Figure 1.** Overview of the NP-KG framework that combines biomedical ontologies, drug databases, natural products, and domain-specific literature information.

### 2.1. Data Integration

An ontology is a formal representation of knowledge in a domain of discourse. Ontologies are used to formally structure the classes and relations within a domain, where classes refer to sets of entities in the world and relations are properties that hold for two or more entities [21]. For example, the Chemical Entities of Biological Interest (ChEBI) is a biomedical ontology representing molecular entities, focusing on chemical compounds and relations between the entities [22]. 'Catechin' (ChEBI:23053) is an example of a class in ChEBI ontology and has a unique identifier, name, and definition. Biomedical ontologies represent our current understanding of biological reality and can be used to represent knowledge as knowledge statements. The Open Biological and Biomedical Ontology (OBO) Foundry is a family of interoperable ontologies in the biomedical domain following a set of principles that include open use, non-overlapping content, and common syntax, as well as relations [23]. As ontologies contain entities in a domain and relations between the entities, they can be represented as graphs, where nodes of the graph represent classes or instances and edges represent an axiom or relation involving the connected classes or instances [21].

We used the Phenotype Knowledge Translator (PheKnowLator) workflow to construct an ontology-grounded KG consisting of the OBO Foundry ontologies and linked data sources. PheKnowLator (v3.0.0) is a Python library that constructs large-scale, biomedical KGs by semantically integrating biomedical ontologies and complex heterogeneous data [24]. Entities from the OBO Foundry ontologies (diseases from Mondo Disease Ontology [25]; phenotypes from Human Phenotype Ontology [26]; anatomical entities from Uber Anatomy Ontology [27]; biological processes, cellular components, and molecular functions from Gene Ontology [28]; proteins from Human Protein Ontology [29]; pathways from Pathway Ontology [30]; chemicals from ChEBI ontology [22]; genes and variants from Sequence Ontology [31]; and cells from Cell Ontology [32] and Cell Line Ontology [33]) and linked data sources were integrated in the KG with the PheKnowLator workflow.

In addition to the ontologies and data sources originally included in the PheKnowLator workflow [34], we extended the workflow to include the Ontology of Adverse Events [35] and data from the following drug data sources in the ontology-grounded KG:

- Drug Interaction Knowledge Base (DIKB) (v2017) [36,37]: evidence of enzyme substrates and inhibition, including *in vitro* information, drug label statements, and results from randomized clinical trials. To simplify the knowledge representation, only the positive evidence present in DIKB was included in the KG.

- Drug Central database (v2017) [38]: *in vitro* evidence of enzyme inhibition, drug-transporter interactions, drug-bacteria interactions, and enzyme and transporter substrates.

- FDA Drug Interaction database (v2017) [39]: *in vitro* and clinical evidence of enzyme and transporter substrates and inhibitors. We included all data where the reported fold change in area under the receiver operating characteristic curve of the drug substrate for an enzyme or transporter is at least 2-fold in the presence of a purported inhibitor drug. This cut-off was chosen because it represents strong positive evidence of a clinically measurable pharmacokinetic mechanism (i.e., the primary drug clearance pathway involves a specific enzyme or transporter that can be inhibited by a drug to a clinically measurable extent).

The 2017 versions of the above databases were included to enable time-slicing in the KG. The time-slicing approach for KG evaluation splits the graph to predict chronologically later links. All entities in the above data sources were mapped to the OBO Foundry ontologies before integration into the KG. We then implemented the Network Transformation for Statistical Learning (OWL-NETS) method in the PheKnowLator workflow with a goal to build a semantically rich KG to facilitate hypothesis generation [40]. The OWL-NETS method decoded all Web Ontology Language (OWL)-encoded classes and axioms into clinically and biologically relevant edges in the graph. To extend the PheKnowLator workflow with the above data sources, we applied the instance-based approach for KG construction, where knowledge statements from the ontologies are added to the KG in the form of subject-relation-object triples without the associated logical definitions. The resulting KG was saved as subject-relation-object triples and as a Python NetworkX multidigraph [41], which is a directed graph containing nodes and edges of multiple types.

### 2.2. Representation of Natural Products

Inclusion of natural products and their chemical constituents in the OBO Foundry is limited. Natural products and their constituents are also limited in drug databases such as Drug Bank and Drug Central, which are commonly used to predict drug-drug interactions or drug-related adverse events. The Drug Ontology [42], FoodOn ontology [43], and the NCBI Taxonomy [44] contain natural products but do not provide information about the pharmacokinetic interactions precipitated by natural products or their constituents. The ChEBI ontology contains some natural product constituents, including catechin and the kratom alkaloid mitragynine, but the coverage of constituents and their metabolites is incomplete. We addressed this gap by integrating extensions designed in our prior work in the ChEBI ontology to include information about natural products and link them to their constituents [45]. In brief, we created semantic representations for natural products using data from the Global Substance Registration System [46], including information about natural product constituents and their metabolites, translated them to logical statements in OWL, integrated the statements into the ChEBI (Lite) ontology, and finally into the KG.

### 2.3. Literature-based Graph

The literature-based graph (Figure 2) was created from full texts from the scientific literature related to green tea and kratom.

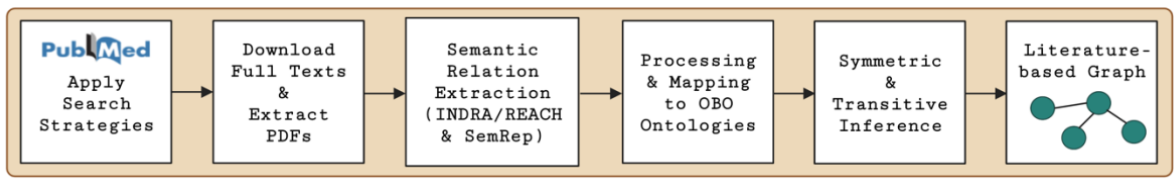

**Figure 2.** Workflow for semantic relation extraction and literature-based graph construction.

### 2.3.1. Semantic Relation Extraction

Semantic relation extraction produces predications from texts. We extracted predications from two biomedical relation extraction systems to add to the ontology-grounded KG. The scope of the literature for the literature-based graph included all PubMed-indexed articles related to green tea and kratom (including keywords of scientific names, synonyms, and their constituents) and pharmacokinetic interactions. The search was restricted to English language articles only and placed no restrictions on animal or human studies. Articles included mechanistic NPDI studies, case reports, and clinical studies from January 1982 to March 2022 for green tea, and from January 1988 to March 2022 for kratom. The search strategies are available in Appendix A.

#### 2.3.1.1. Relation Extraction Systems

SemRep [47] and the Integrated Network and Dynamic Reasoning Assembler (INDRA) [48] with the Reading and Assembling Contextual and Holistic Mechanisms from Text (REACH) biological reader [49] were used to extract semantic relations. SemRep (v1.8) is a natural language processing system that infers relations between entities in biomedical texts and is informed by syntactic and semantic constraints combined with biomedical domain knowledge [47]. SemRep uses MetaMap (v2018) to identify biomedical entities in texts and maps to Unified Medical Language System (UMLS) concepts [50]. SemRep achieved 0.69 precision and 0.42 recall for relation extraction from PubMed titles and abstracts in a test collection [47].

REACH (v1.6.3) is an information extraction system designed to robustly parse full texts of biomedical literature and extract cancer signaling pathways. REACH focuses on extracting information about biochemical interactions of proteins and has demonstrated high precision and throughput in uncovering mechanistic knowledge from the literature compared to manually curated databases and other biological relation extraction systems [49]. Evaluation of REACH for extracting mechanistic information from the biomedical literature showed a precision of 0.62 and estimated recall of 0.486 based on a throughput evaluation [49]. INDRA (v1.19) provides a framework to extract predications using REACH.

We extracted full texts from the literature in our search strategy from PubMed Central and from text mining of PDFs using the PDFMiner library [51] and Python script adapted from Hoang et. al. [52]. Although SemRep has mainly been used for titles and abstracts, our pipeline extended SemRep to process full texts. The available full texts were processed by SemRep and INDRA/REACH to extract predications.

### 2.3.2. Literature-based Graph Construction

To create the literature-based graph from the extracted predications, we linked all subjects, objects, and relations from the predications to the OBO Foundry ontologies to avoid ambiguity and standardize the graph. This process, termed entity linking, can be achieved with automated methods

(such as named entity recognition tools) or curated mappings. We used a combination of both approaches to standardize subjects, objects, and relations in the predications to OBO Foundry ontologies. First, relations extracted by SemRep and INDRA/REACH were standardized to Relation Ontology terms based on domain, range, and definitions of the relations after consulting with experts in the field. SemRep also extracted negated predications involving negative relations such as *neg_inhibits* and *neg_interacts_with*. These were stored separately as there were no corresponding negative relations in the OBO Foundry ontologies. Next, gene functions and phenotype concepts identified by SemRep were mapped to concepts in the Gene Ontology and the Human Phenotype Ontology using database cross references in UMLS. Subjects and objects extracted by the INDRA/REACH system were mapped to the OBO Foundry ontologies using the INDRA BioOntology module.

We next applied the OntoRunNER OGER++ wrapper [53] with custom term lists containing OBO Foundry ontologies (from the ontology-grounded KG) to identify candidate mappings for the remaining unmapped subjects and objects. All candidate mappings were manually reviewed to determine the correct identifiers in the OBO Foundry. Predications with unmapped subjects and objects were excluded. We further filtered the predications to exclude relations that were not useful for hypothesis generation, such as *converts_to, diagnoses, method_of,* and *process_of*. Certain semantic types (such as activities and behaviors, concepts and ideas, organizations) and generic concepts (such as Animals, Disease, Persons, Organism, Syndrome, Patients) provided by Zhang et. al. [54] were also excluded. We used the INDRA assembly module for deduplicating, normalizing namespaces, and generating confidence scores for the predications extracted from INDRA/REACH.

We then applied symmetric and transitive closure over the predications using the CLIPSPy library [55]. The closure process used entailment properties to infer new predications (termed inferred predications) based on the rules for symmetry and transitivity of the relations. We included the relations *'interacts_with'* and *'molecularly_interacts_with'* in the symmetric closure rules and the relations *'part of'*, *'precedes'*, and *'positively_regulates'* in the transitive closure rules based on the relation properties in the Relation Ontology. As an example, applying closure over a predication involving a symmetric relation, such as '*catechin interacts_with nadolol*', results in the inferred predication '*nadolol interacts_with catechin*'. The inferred predications were combined with those extracted from the literature. We then created the literature-based graph as a NetworkX graph (separate from the ontology-grounded KG) from the combined predications, where the nodes in the graph represented the subjects and objects from the predications, and relations formed the edges between the nodes. The metadata related to each predication, including the PubMed ID of the article, year of publication of the article, confidence score (if available), and the sentence from which the predication was extracted, was also added in the graph.

### 2.4. Evaluation

We combined the ontology-grounded KG and the literature-based graph to create NP-KG and store as another NetworkX multidigraph. This combined KG is referred to as NP-KG in the following sections. We evaluated NP-KG with case studies of green tea and kratom. We also conducted an error analysis to identify errors from relation extraction and contradictory information in NP-KG.

#### 2.4.1. Ground Truth

Potential pharmacokinetic interactions that have been identified through *in vitro* experiments and clinical studies were considered as ground truth for NPDIs. If a clinical study had not yet been conducted, results from *in vitro* experiments and mechanistic static model predictions sufficed as ground truth. In case of contradictory results between *in vitro* and clinical studies, the clinical studies were considered the ground truth. We obtained ground truth for KG evaluation from the Center of Excellence for Natural Product Drug Interaction Research (NaPDI Center). The NaPDI Center maintains a human curated database that contains results from *in vitro* experiments and clinical pharmacokinetic NPDI studies [6].

#### 2.4.2. Evaluation Strategies

##### I. Knowledge Recapturing

We evaluated the potential of NP-KG to capture existing knowledge about green tea and kratom interactions with enzymes and transporters. We first collected OBO Foundry ontology identifiers for the enzymes and transporters in the NaPDI Center database. Next, for all nodes related to green tea (green tea and the green tea catechins (-)-epicatechin-3 gallate (ECG), (-)-epigallocatechin gallate (EGCG), epicatechin, catechin, and gallocatechin) and kratom (kratom and the kratom alkaloid mitragynine), we extracted direct edges (if any) and shortest paths (if any) from the nodes to enzymes and transporters in NP-KG. Edges and paths extracted from NP-KG were compared to the ground truth information in the database. Results from *in vitro* experiments and the published literature in the database involving inhibition, induction, and no interactions for the enzymes and transporters were included in the comparison.

We calculated the number of congruences, contradictions, and missing knowledge from NP-KG while comparing to data in the database. Congruence was established when the direct edge or shortest path in NP-KG concurred with the ground truth information about the interaction. Contradiction was established if the direct edge or shortest path in NP-KG contradicted the results of the interaction in the database. An example of congruence and contradiction between nodes X and Y is shown (Figure 3). In case of both congruence and contradiction, we manually reviewed the evidence related to the information from metadata present in NP-KG. For edges or paths where congruence or contradiction could not be established, such as edges or paths involving the relations *interacts_with*, *molecularly_interacts_with*, and *directly_regulates_activity_of*, the edges and paths were not included in the calculation of congruences or contradictions. Shortest paths were obtained using the bidirectional shortest path algorithm in the NetworkX library.

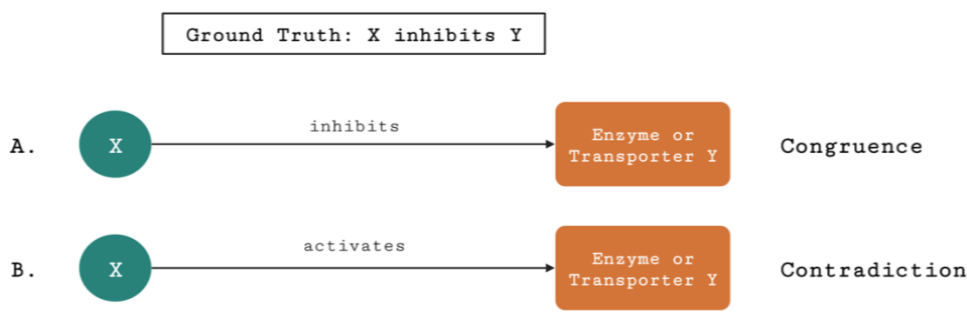

**Figure 3.** Congruence and contradiction between node X and an enzyme or transporter Y in NP-KG, where ground truth information between the nodes states that X inhibits Y.

##### II. Meta-path Discovery

Meta-paths are defined as sequences of connected relations between nodes, node types, or sequences of node types in a KG. They are commonly used in KG construction and hypothesis generation

[54,56]. We applied direct edge searches and meta-paths in NP-KG to generate hypotheses for the pharmacokinetic NPDIs involving green tea and kratom. Figure 4 shows the direct edges and meta-path searches applied for discovering the interactions between a natural product or its constituent and a drug, with an interacting enzyme or transporter. We selected 5 natural product-drug pairs for the meta-path discovery, namely green tea-raloxifene, green tea-nadolol, kratom-midazolam, kratom-quetiapine, and kratom-venlafaxine. These interactions were chosen due to the published clinical results and case reports that provided evidence and potential mechanisms for the interactions [8,15,16,57–60].

Relations A and B in the direct edges and meta-path definitions (Figure 4) were filtered to include the relations *interacts_with, molecularly_interacts with, associated_with, directly_regulates_activity_of, positively_regulates, inhibits, capable_of_regulating, capable_of_positively_regulating, correlated_with, is_substrate_of, transports,* and *regulates_activity_of*. We then evaluated the results based on the available ground truth information for the interactions.

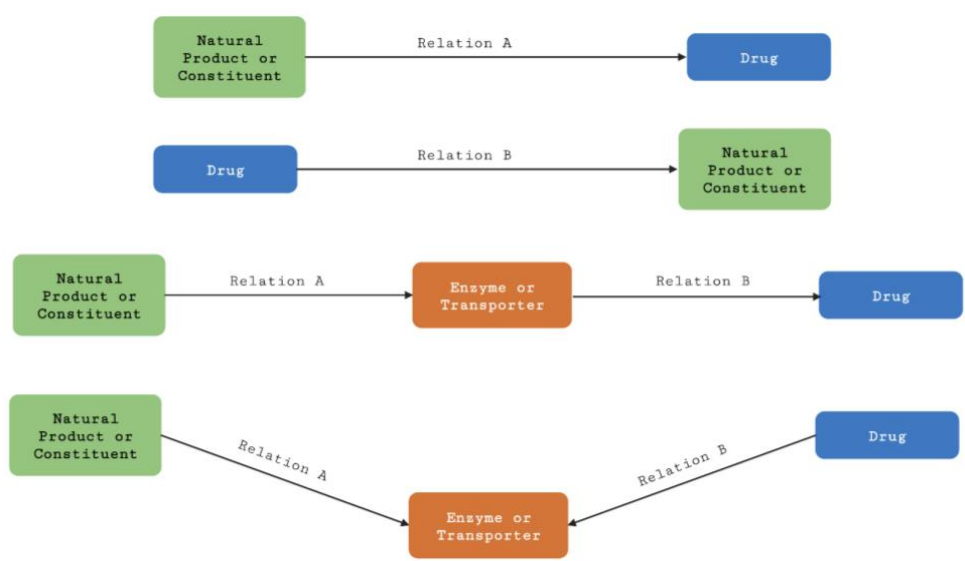

**Figure 4.** Direct edges and meta-paths for natural products or natural product constituents and drugs with an interacting enzyme or transporter.

### 3. Results

#### 3.1. Data Integration

The ontology-grounded KG created from the PheKnowLator workflow included entities from ontologies representing diseases, genes, phenotypes, chemicals, anatomy, cell lines, proteins, and pathways from 10 OBO Foundry ontologies and 13 publicly available data sources [34]. After integrating green tea, kratom, and their constituents and metabolites, 15 classes, 96 axioms, and 13 individuals were added to ChEBI Lite ontology, bringing the total to 156,113 classes, 1,201,077 logical axioms, and 13 individuals. The merged ChEBI ontology, individual semantic representations, and logical extensions are publicly available [45].

The ontology-grounded KG contained 745,250 nodes and 7,224,186 edges. Appendix A Table A.1 shows the number of edges in the ontology-grounded KG with the edge type, edge source, and edge label. The edges *chemical-transporter* (N=90)*, chemical-molecule* (N=391)*, chemical-inhibitor*

(N=272), and *chemical-substrate* (N=514) were included in the extended workflow from the drug data sources.

## 3.2. Literature-based Graph

We obtained full texts of 735 scientific articles for green tea and 59 articles for kratom from PubMed. INDRA/REACH extracted 12,270 predications from 716 full-text articles (89.05%). SemRep extracted 250,360 predications from 637 full-text articles (79.23%). There were 13,676 predications from SemRep for green tea and kratom after deduplication, concept reduction, and filtering. There were 7,999 predications for green tea and kratom from INDRA/REACH after processing with the INDRA assembly module. Implementation of the transitive and symmetric closures over the predications resulted in an additional 13,637 inferred predications. Appendix Table A.2 describes the predications extracted from the literature with their counts, source, and the OBO Foundry ontology mapped relation.

The literature-based graph constructed from the combined and deduplicated predications for green tea and kratom contained 3,510 nodes and 25,421 edges. Figure 5 presents an overview of the types of nodes and edges in the literature-based graph. Figure 6 shows the distribution of the edges in the literature-based graph by edge type.

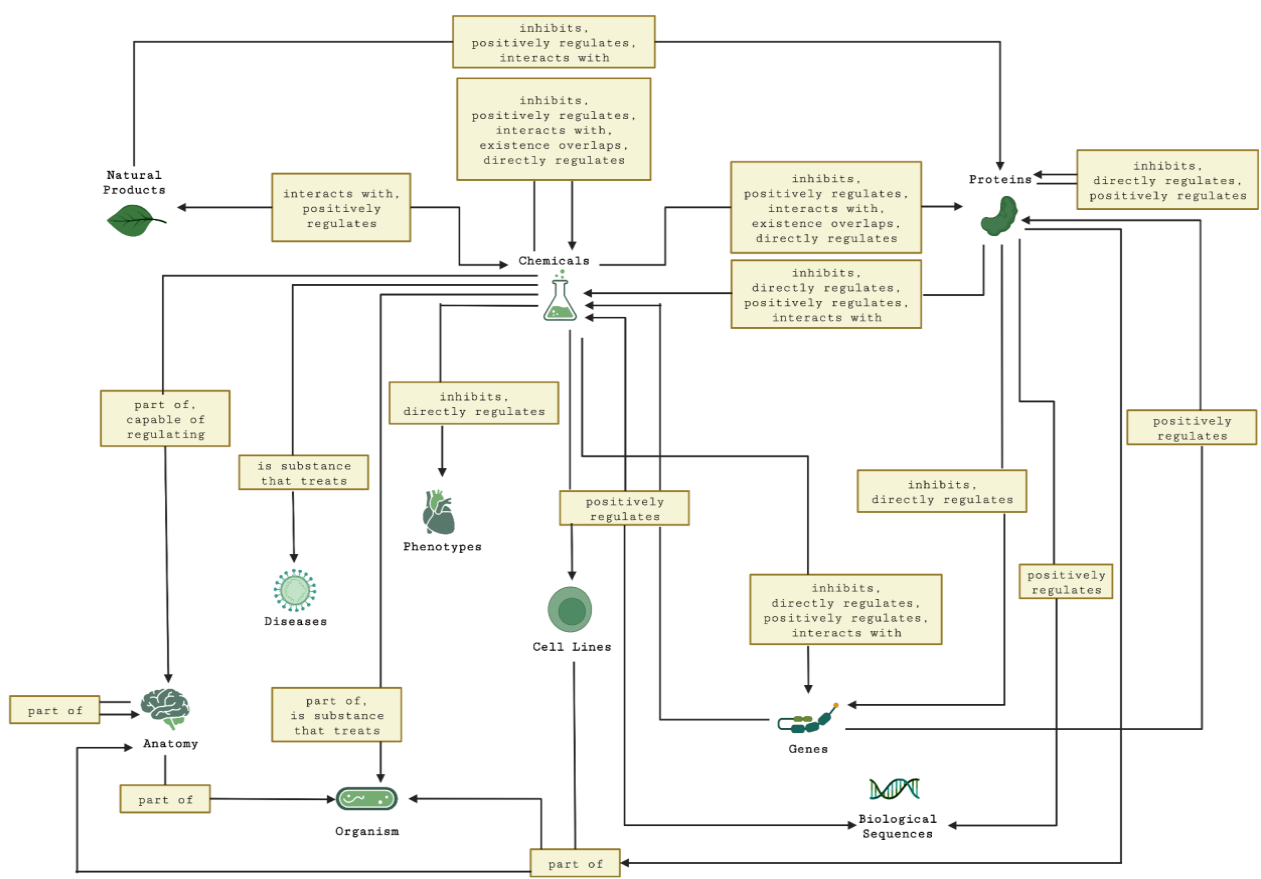

**Figure 5.** Node types and edge types in the literature-based graph (with edge counts greater than 100). Rectangles represent edges with edge labels from the graph.

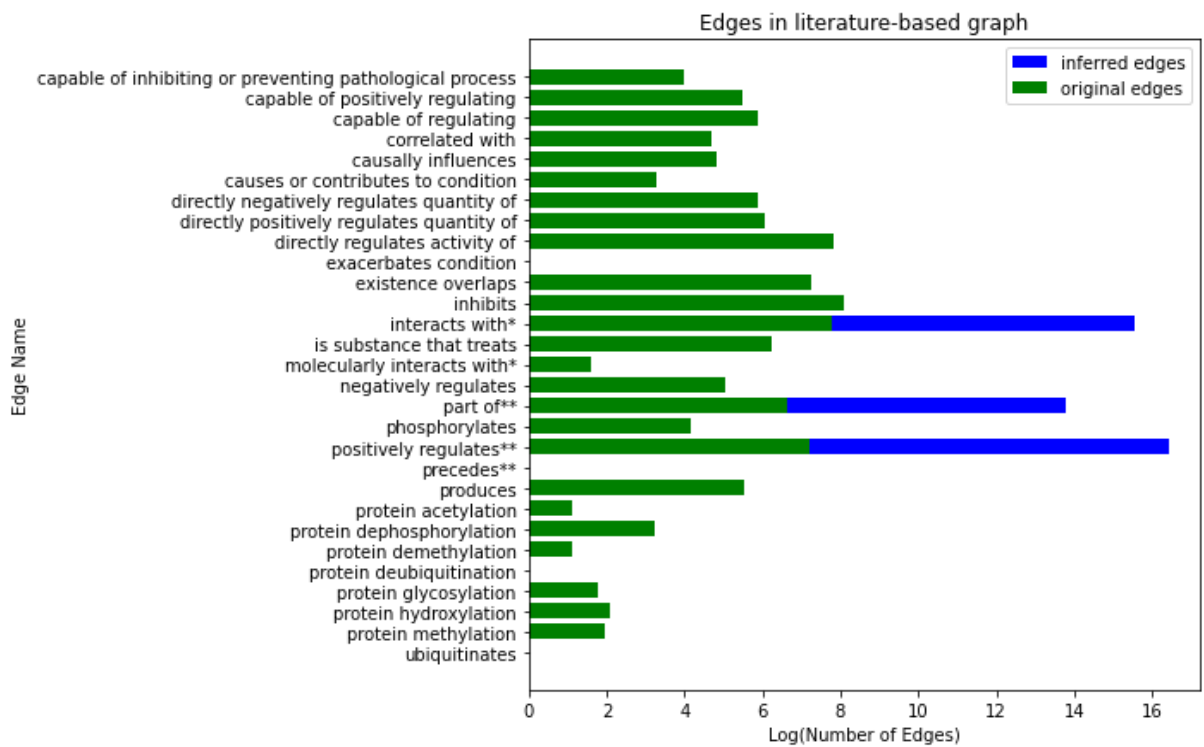

**Figure 6.** Distribution of edges in the literature-based graph, including edges extracted from the literature and inferred edges. The x-axis presents the log of the number of edges and y-axis presents the edge labels mapped to OBO Foundry ontologies. Edges included in the symmetric closure are marked with *, and edges included in the transitive closure are marked with **.

### 3.3. Evaluation

After integration of the ontology-grounded KG and the literature-based graph, the heterogeneous NP-KG framework contained 745,512 nodes and 7,249,576 edges. The literature-based graph added 262 unique nodes and 25,390 unique edges to NP-KG. Table 1 describes the graph counts for the ontology-grounded KG, the literature-based graph, and NP-KG.

**Table 1.** Graph counts for ontology-grounded KG, literature-based graph, and NP-KG

| Parameter | Ontology-grounded KG | Literature-based Graph | NP-KG (% change from ontology-grounded KG) |
| --- | --- | --- | --- |
| **Nodes** | 745,250 | 4,157 | 745,512 (0.035%) |
| **Edges** | 7,224,186 | 27,784 | 7,249,576 (0.35%) |
| **Average Degree** | 9.69 | 6.68 | 9.72 (0.31%) |
| **Node Density** | 1.301 x 10$^{-5}$ | 0.002 | 1.304 x 10$^{-5}$ (0.23%) |

I. Knowledge Recapturing

Table 2 summarizes the congruent and contradictory knowledge from NP-KG related to pharmacokinetic interactions between the natural products-related nodes and enzymes and transporters when compared to the ground truth information. For the green tea-related nodes, we performed 59 searches for direct edges or shortest paths in NP-KG involving 19 enzymes and 8 transporters. For the kratom-related nodes, we performed 14 searches for direct edges or shortest paths involving 10 enzymes and 1 transporter. Figure 7 shows examples of direct edges and shortest paths in NP-KG. Some nodes contained multiple edges between them, such as catechin and UGT1A1 (Figure 7a), mitragynine and cytochrome P450 (CYP) 3A (Figure 7d), and mitragynine and CYP2D6 (Figure 7e). Results with both congruent and contradictory edges between the nodes were manually reviewed to verify congruence and/or contradiction.

**Table 2.** Summary of congruences and contradictions for direct edges and shortest paths in NP-KG compared to ground truth information.

|  | **Green Tea (%)** | **Kratom (%)** |
|---|---|---|
| **Congruence** | 23 (38.98) | 7 (50.0) |
| **Contradiction** | 9 (15.25) | 3 (21.43) |
| **Edges/paths exist but no congruence or contradiction** | 25 (42.37) | 3 (21.43) |
| **Both congruence and contradiction** | 2 (3.39) | 1 (7.14) |
| **Total searches** | 59 | 14 |

Examples of manually reviewed searches include the following:

- The edges *inhibits, positively_regulates,* and *directly_positively_regulates_quantity_of* between EGCG and CYP1A2 and EGCG and UGT1A1 suggest both congruence and contradiction. Manual review showed that contradictions in the literature were responsible for the evidence of both congruence and contradictions in NP-KG.
- Mitragynine, the major alkaloid in kratom leaves, is a time-dependent inhibitor of CYP3A and reversible inhibitor of CYP2D6 [58]. The edges *inhibits* and *directly_positively_regulates_quantity_of* between mitragynine and CYP3A4 (Figure 7d) imply both congruence and contradiction with the ground truth. Manual review of the edges shows evidence of statements in the literature suggesting mitragynine inhibits (PubMed ID: 24174816) and induces (PubMed ID: 23274770) CYP3A4 *in vitro*.
- The *inhibits* edges and *positively_regulates* edge between mitragynine and CYP2D6 (Figure 7e) imply both congruence and contradiction with ground truth information. Manual review in this case revealed errors from relation extraction for the *positively_regulates* edge between mitragynine and CYP2D6.

Congruence or contradiction could not be established for 25 (42.37%) searches for green tea nodes and 3 (21.43%) searches for kratom nodes as the edges included general interaction predicates such as *interacts_with* and *regulates_activity_of*. For example, NP-KG found that *mitragynine interacts_with CYP2C19*, although the ground truth is that *mitragynine inhibits CYP2C19*. The search was thus insufficient to conclude congruence or contradiction in this case. Supplementary File 1 presents edges and paths between nodes for the knowledge recapturing evaluation strategy for all kratom and green tea-related nodes.

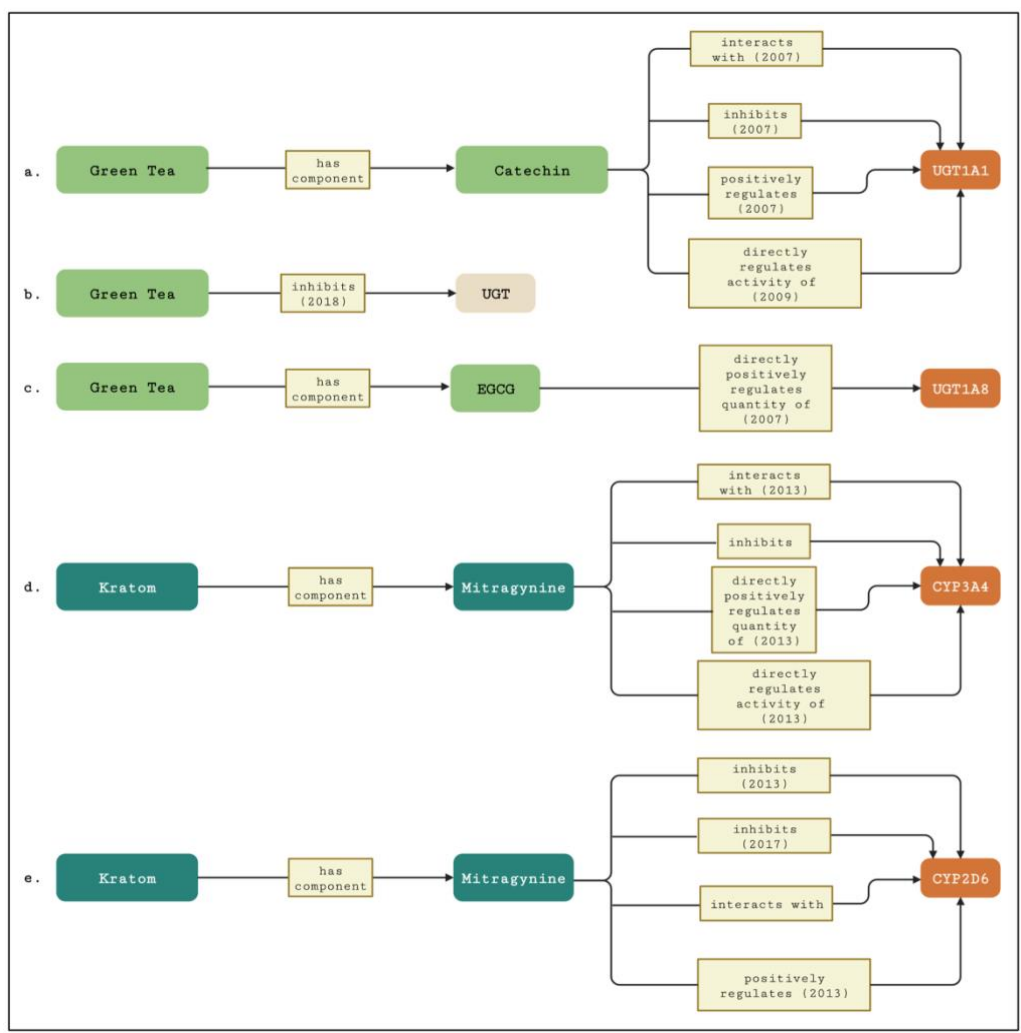

**Figure 7.** (a-c) Edges and/or shortest path(s) between green tea and UGT enzymes (UGT, UGT1A1, UGT1A8, and UGT1A10) in NP-KG. (d-e) Shortest path(s) between kratom and cytochrome P450 enzymes (CYP3A4, CYP2D6) in NP-KG. If an edge in the path is derived from the literature, the year of publication is noted with the edge label. The rounded rectangles represent nodes and rectangles represent edges in NP-KG.

### II. Meta-path Discovery

We applied direct edge and meta-path searches for discovery of pharmacokinetic NPDIs involving green tea and kratom in NP-KG (Figure 4). Table 3 summarizes the results from the direct edges and meta-path discovery for green tea-raloxifene, green tea-nadolol, kratom-midazolam, kratom-quetiapine, and kratom-venlafaxine.

**Table 3.** Summary of results of meta-path discovery in NP-KG for the natural product-drug pairs. (ND=None Detected)

| Natural product-drug pair | Enzyme(s) | Transporter(s) |
|---|---|---|
| **Green tea-raloxifene** | CYP3A4, UGT | ND |
| **Green tea-nadolol** | See Appendix Table A.4 | See Appendix Table A.5 |
| **Kratom-midazolam** | CYP2D6, CYP3A4 | P-glycoprotein |
| **Kratom-quetiapine** | CYP3A4 | P-glycoprotein |
| **Kratom-venlafaxine** | CYP2D6, CYP3A4 | ND |

*Green tea-raloxifene*: Raloxifene is an anti-osteoporosis agent and an intestinal UGT substrate. Results from *in vitro* experiments showed concentration-dependent inhibition of intestinal UGT activity by green tea and green tea catechins [15]. In the subsequent clinical study conducted to determine the effects of green tea on raloxifene pharmacokinetics in 16 healthy adult participants, an interaction was observed. Specifically, relative to water, green tea decreased the area under the plasma concentration *versus* time curve (AUC) of raloxifene by ~30%. However, the change in raloxifene AUC was opposite to the IVIVE prediction (increase in raloxifene AUC). These observations, combined with the lack of effect of green tea on raloxifene half-life and the raloxifene-to-glucuronide AUC ratios, suggested that inhibition of intestinal UGT activity was not responsible for the interaction [16]. Meta-path discovery in NP-KG identified CYP3A4 and UGTs for the pharmacokinetic interaction between green tea and raloxifene. NP-KG also identified raloxifene to be a substrate for CYP2C9 but with no connection to green tea. Figure 8 shows a graphical snapshot of the paths among green tea, green tea constituents, and raloxifene nodes in NP-KG focusing on pharmacokinetic interactions.

*Green tea-nadolol*: Nadolol is a non-selective beta-adrenoceptor blocker used to treat hypertension. A clinical pharmacokinetic interaction study involving a green tea beverage and nadolol showed green tea to decrease nadolol AUC by 80% and to decrease the systolic blood pressure lowering effects of nadolol in ten healthy adults [8]. The *in vitro* experiments suggested that nadolol is a substrate for OATP1A2 and EGCG inhibited OATP1A2-mediated nadolol uptake into OATP1A2-expressing human embryonic kidney cells. However, this mechanism has not been confirmed as the existence of OATP1A2 is equivocal, and the interaction may involve other mechanisms, such as upregulation of efflux transporters [8]. Meta-path discovery in NP-KG identified 8 drug-metabolizing enzymes and 12 transporters for the pharmacokinetic interaction between green tea and nadolol. The list of drug metabolizing enzymes and transporters is available in Appendix Tables A.4 and A.5 respectively.

*Kratom-midazolam*: Midazolam is a short-acting benzodiazepine and a CYP3A substrate. A kratom extract and the kratom alkaloid mitragynine have been shown to inhibit CYP2D6, CYP2C9, and CYP3A activities *in vitro* [58,59]. A mechanistic static model predicted a pharmacokinetic interaction between kratom and midazolam based on time-dependent inhibition of CYP3A by mitragynine [58]. Meta-path discovery in NP-KG identified 2 interacting enzymes, CYP3A4 and CYP2D6, and one transporter, P-glycoprotein, for the pharmacokinetic interaction between kratom and midazolam. However, midazolam is not a substrate for CYP2D6 or P-glycoprotein

*Kratom-quetiapine and kratom-venlafaxine*: Venlafaxine is an antidepressant and a dual CYP3A/CYP2D6 substrate. Quetiapine is an antipsychotic and a CYP3A substrate. A recent case report described potential pharmacokinetic kratom-drug interactions involving these two drugs [60]. Proposed mechanisms for the interactions include inhibition of CYP3A and CYP2D6 [60]. Meta-path discovery in NP-KG identified the interacting enzyme, CYP3A4, and the transporter, P-glycoprotein, for the interaction between kratom and quetiapine, and the interacting enzymes, CYP3A4 and CYP2D6, for the interaction between kratom and venlafaxine. Figures 9a and 9b show graphical snapshots for paths between kratom and interacting drugs.

Appendix Table A.3 presents the results from the meta-path discovery searches for the natural product-drug pairs examined.

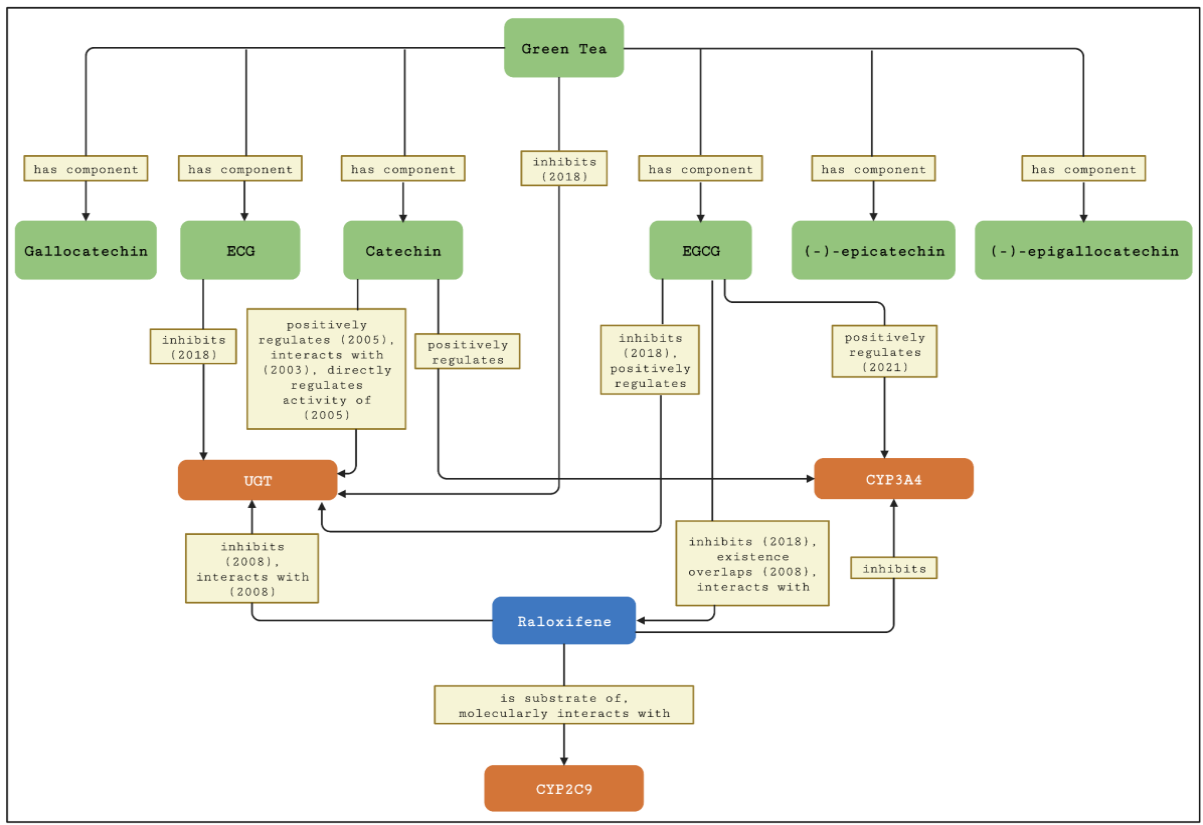

**Figure 8.** Nodes and edges in NP-KG related to pharmacokinetic interactions between green tea and raloxifene with enzymes (CYP3A4, CYP2C9, UGT). Rounded rectangles represent nodes and rectangles represent edges in NP-KG. If an edge is derived from literature, the year of publication is noted with the edge label. Note that nodes can contain multiple edges between them.

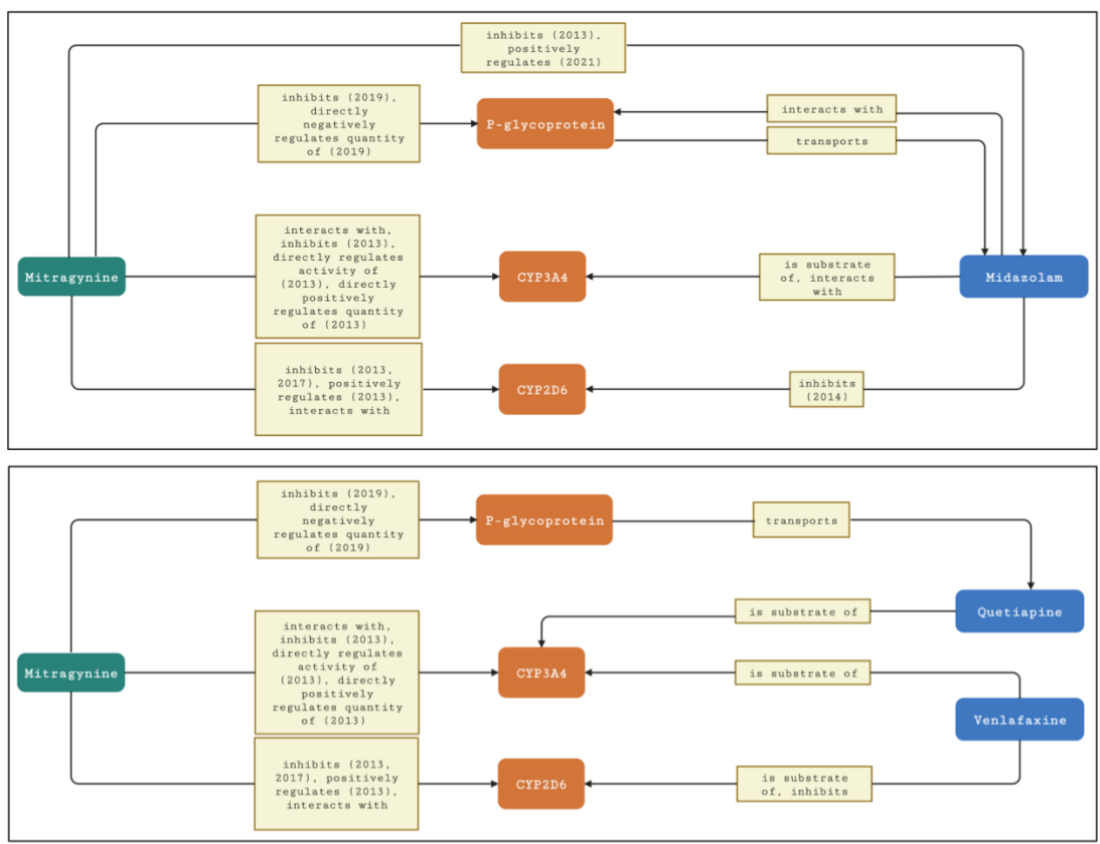

**Figure 9a.** Nodes and edges in NP-KG related to pharmacokinetic interactions between mitragynine and midazolam with interacting enzymes (CYP3A4, CYP2D6) and a transporter (P-glycoprotein). **9b.** Nodes and edges in NP-KG related to pharmacokinetic interactions between mitragynine and drugs (quetiapine and venlafaxine) with interacting enzymes (CYP3A4, CYP2D6) and a transporter (P-glycoprotein). Rounded rectangles represent nodes and rectangles represent edges in NP-KG. If an edge is derived from the literature, the year of publication is noted with the edge label.

**Error Analysis**

Error analysis of predications related to green tea-drug interactions (Figures 7-8) and kratom-drug interactions (Figures 7, 9a-b) showed that multiple incorrect predications exist in NP-KG generated by the relation extraction systems. Examples of incorrect predications included *midazolam inhibits CYP2D6 (2014), mitragynine inhibits UGT2B7 (2014), mitragynine stimulates midazolam (2021), 7-hydroxymitragynine inhibits mitragynine (2014)*.

Manual verification of the predications identified broad errors from the relation extraction systems from the pharmacokinetic literature. First, sentences in texts containing 'increase or decrease in area under the plasma concentration-time curve' were commonly misread by the systems and not represented accurately by the extracted predication. Second, in some cases, negation in sentences was not accurately extracted, leading to inaccurate predications. Third, substrates of enzymes and transporters mentioned in the literature are not extracted by either of the relation extraction systems as they do not contain rules to identify the 'substrate of' relation in text. Improving relation extraction from the literature and applying constraints to NP-KG during construction and hypothesis generation can help resolve some of the errors. Other errors from the ontology-grounded KG also exist, such as *midazolam interacts_with CYP2E1* (source: Comparative Toxicogenomics Database), *UGT1A4*

*molecular_interacts_with UGT1A6* (source: STRING Database)*, and CYP1A2 molecularly_interacts_with CYP2C8* (source: STRING Database).

There are several contradictory edge pairs between nodes in NP-KG that are not included in the errors described above. Contradictory edges may be defined as edges that contradict each other semantically, such as *inhibits* and *positively_regulates*. For example, the pairs *mitragynine inhibits CYP2D6 (2013, 2017)* and *mitragynine positively_regulates CYP2D6 (2013)*, *mitragynine inhibits CYP3A4 (2013)* and *mitragynine directly_positively_regulates_quantity_of CYP3A4 (2013), EGCG inhibits UGT (2018)* and *EGCG positively_regulates UGT* are contradictory. These contradictions can be reasonable in cases where the evolution of knowledge leads to multiple edges between nodes, as the evidence related to the nodes is updated (in the literature or elsewhere). Although we can track the year of publications for the sources of the predications, manual verification and fact-checking of the contradictory edges are required to improve the hypotheses generated by NP-KG. These procedures can be aided by automated filtering strategies for incorrect predications due to relation extraction [61].

### 5. Discussion

To our knowledge, this study is the first to combine existing curated information from biomedical ontologies with computable knowledge extracted from full texts of scientific articles to construct a KG focused on NPDIs. NP-KG is a novel computational framework that combines biomedical ontologies, publicly available databases, and domain-specific information from the scientific literature to generate mechanistic hypotheses for potential pharmacokinetic NPDIs. We demonstrated that NP-KG can be used to recapture knowledge about pharmacokinetic NPDIs, as well as provide a template for predicting NPDIs that have not been studied extensively to guide researchers in the design and conduct of new studies.

We applied two evaluation strategies to validate the hypotheses generated from NP-KG for green tea- and kratom-drug interactions. The first provided direct edges and shortest paths between nodes related to the natural products in NP-KG and potential interacting enzymes and transporters. When compared to curated data from the NaPDI Center database, we identified both congruent and contradictory information in NP-KG. Further analysis showed that contradictory information was the result of contradictions present in the literature or errors from the relation extraction systems we used. Manual review was required in multiple cases where both congruent and contradictory edges existed between the nodes (15.25% searches for green tea, 21.43% searches for kratom). Although our targeted searches could identify some contradictions, a complete contradiction analysis of NP-KG was out of scope for this study.

Resolving contradictions in literature-based discovery and biomedical KGs is an active area of research. An investigation of clinically relevant, contradictory predications in SemMedDB [62] (semantic predications database for PubMed abstracts generated using SemRep) revealed multiple reasons for the contradictions, including characteristics related to the patients, known controversies, contradictions present in the literature, and the progression of knowledge [63]. Although the analysis was restricted to abstracts in the clinical domain, both the immediate context (source sentence) and full abstract were required to detect and resolve the contradictions [63]. Similarly, Sosa and Altman [64] demonstrated that literature-based KGs suffer from several challenges that give rise to contradictory information in the drug repurposing domain, namely, errors in relation extraction, errors in entity normalization leading to false positive contradiction pairs, contradictions in the biomedical literature due to progression of science and variation in research methods, and inconclusive evidence. Enhancing NP-KG with contextual information, including more granular information, quantitative facts (e.g., dosage), negative results, confidence scores, and source sections of the predications can help resolve contradictions and improve knowledge inference.

The second evaluation strategy for NP-KG applied direct edge and meta-path searches for select natural product-drug pairs, specifically green tea-raloxifene, green tea-nadolol, kratom-midazolam, kratom-quetiapine, and kratom-venlafaxine. Meta-paths are also known as discovery patterns or pathways in literature-based discovery and can be of varying lengths. Prior work suggests that they can correspond to mechanisms of pharmacological efficacy and thus may be used to generate plausible hypotheses from KGs [56]. In related work, Schutte et. al. [11] applied meta-paths (*Dietary_Supplement-Gene-Drug* and *Dietary_supplement-Gene1-Biological_Function-Gene2-Drug*) to identify NPDIs involving supplements from a literature-based KG. Meta-path discovery for the green tea-raloxifene interaction identified CYP3A4 and UGTs involved in the mechanism. Although inhibition of intestinal UGT by green tea catechins is supported by *in vitro* data, the clinical study did not confirm this mechanism [15,16]. This *in vitro-in vivo* disconnect suggests that NP-KG may need to be expanded to more data domains, including physicochemical properties of drugs and natural products, and the gut microbiome domain such as with the gutMEGA database [65] and microbe-drug associations from the microbiota-active substance interactions database [66], to explore potential mechanistic hypotheses.

For the green tea-nadolol interaction, although the meta-path searches identified OATP1A2 as one of the transporters involved in the interaction, other enzymes and transporters may or may not be involved in the interaction and require further review. There is no evidence to support that nadolol is a substrate of the CYP enzymes identified by NP-KG for the interaction between green tea and nadolol (Appendix Table A.4). NP-KG meta-path discovery results suggested that mechanisms of putative pharmacokinetic interactions between kratom and midazolam, quetiapine, and venlafaxine primarily involve CYP3A4, CYP2D6, and P-glycoprotein. Some of these results are supported by the published literature [58,67–69] while some are yet unknown. The evaluation results suggest that NP-KG provides a major advance in support of our hypothesis for computational discovery of pharmacokinetic NPDI mechanisms by combining biomedical ontologies and databases with the scientific literature.

Constructing a KG by integrating full texts of the scientific literature with biomedical ontologies is challenging due to the limited availability of full texts in computable format and lack of interoperability within vocabularies used in ontologies, databases, and relation extraction systems. Large-scale KGs developed for drug-related applications, such as drug-drug interaction, drug-target prediction, and drug repurposing [54,56,70,71], are typically constructed either from curated data sources (such as ontologies, drug databases) or the scientific literature. KGs constructed from the scientific literature typically rely on manual annotations or automated relation extraction systems (e.g., SemRep). The literature-based graph constructed in this study combined predications from two relation extraction systems for more than 700 full-text articles. Node and edge types in the literature-based graph (Figures 5 and 6) showed that the relation extraction systems can capture relevant information from the scientific literature for different types of biomedical entities and supplement the ontology-grounded KG with domain-specific knowledge.

Automated extraction and integration of the predications from the literature enabled us to draw inferences from NP-KG suggested in the published literature, and metadata included in the graph were used to verify the accuracy of the edges. Data integration and KG inference were further supported by the choice of instance-based representation for data integration, where subject-relation-object triples were retained in the ontology-grounded KG and the literature-based graph. The standardization of relations extracted by SemRep and INDRA/REACH to OBO Foundry ontologies presented in this study (Appendix Table A.2) will further facilitate the integration of literature-based knowledge with biomedical ontologies. Analysis of the predications also revealed errors in the relation extraction systems that, if improved, can lead to more accurate results and fewer contradictions.

Our overall goal is to use NP-KG to understand the knowledge gaps pertaining to NPDIs and generate mechanistic hypotheses for novel NPDIs. To generalize the above approach for all NPDIs, we need broader inclusion of knowledge domains to represent the mechanisms, broader processing of published literature, improved capture of context to address contradictions and assign confidence to statements, and rigorous evaluation of inference relative to specific use cases. Rigorous evaluation would also involve a comprehensive time-slicing approach for hypothesis generation. Upon further improvement in NP-KG based on the gaps identified in our evaluation strategies, NP-KG can be used to generate hypotheses and verify biological plausibility of pharmacovigilance signals from spontaneous reporting systems. Recent studies to detect adverse events related to natural products, and NPDIs from spontaneous reporting systems, including the FDA Adverse Event Reporting System and CFSAN Adverse Event Reporting System, have shown promise [72–74]. The hypotheses can then be validated through manual review by domain experts, *in vitro* experiments, physiologically based pharmacokinetic modeling and simulation, and clinical studies [75,76]. For example, NP-KG can generate hypotheses for known and novel NPDIs with supporting evidence, which can be checked by scientists to validate or refute the hypotheses without conducting a full literature review of known evidence. This approach would facilitate human-machine collaboration for understanding mechanisms of novel NPDIs to better balance the risks and benefits of the concomitant use of natural products and pharmaceutical drugs.

**Limitations and Challenges**

There are several limitations to this study. First, no manual filtering was used for the search strategies, which could have added non-relevant predications to the literature-based graph. Second, the literature-based graph contains known relation extraction errors due to incorrect entity recognition, missing predications (due to low recall), and complex sentence syntax [49]. The reported recall for relation extraction from abstracts with SemRep is 42% [47]; however, our preliminary evaluation showed that the recall (31%) is lower for full texts from the pharmacokinetic literature compared to human curated data (results presented in a poster presentation). SemRep has not been formally evaluated on full texts. Third, the relation extraction systems extracted predications from the full texts without targeting specific sections or information in tables. Fourth, our current approach for entity linking in the literature-based predications mapped only 60-70% of all subjects and objects in the predications, leading to potential missing information in NP-KG. Extensions of named entity recognition systems with custom data sources and relation extraction systems for natural products, such as in SemRepDS [11], can help improve entity linking in the KG. Potential data sources include the Integrated Dietary Supplements Knowledge Base [77] and FoodKG [78] to increase representation of natural products in existing resources. Fifth, searching with meta-paths and shortest paths restricts the search space in the KG, and although this approach is computationally efficient, results are limited. For example, meta-paths applied in this study do not use information about metabolites of the natural product or the drug that could improve the understanding of the mechanism(s) underlying a given NPDI. Finally, validation of NPDI mechanisms is challenging due to the limited availability of ground truth information. Drug Bank and Drug Central do not include data on NPDIs that can be used to evaluate predictions, as is common for drug-drug interactions. The Natural Medicine Interaction Checker database and Natural Medicines Comprehensive Database provide limited information for validating new hypotheses for drug-supplement interaction mechanisms [10,11]. Validation of NPDIs is thus either accomplished through the published literature or adaptation of drug-drug interaction datasets for food-drug interactions [10–12,79]. Our evaluation focused on green tea and kratom with ground truth information extracted from the NaPDI Center database. We only searched for enzymes or transporters with known interactions in the knowledge recapturing evaluation strategy. A broader search for other pharmacokinetic targets (e.g., nuclear receptors) may lead to more insights into these complex interactions.

**6. Conclusions**

NP-KG combined existing curated information in the form of ontologies and databases with domain-specific full texts of scientific articles to construct a KG focused on NPDIs. We demonstrated the potential of NP-KG to generate hypotheses that are grounded in both existing knowledge and suggested mechanisms from the scientific literature. NP-KG can be used to narrow the scope of potential hypotheses for pharmacokinetic NPDIs and help scientists in the preclinical assessment of potential clinically significant NPDIs. Future work will seek to address limitations of the relation extraction systems, extend NP-KG to other natural products, and apply improved hypothesis generation methods and evaluation. We are adding more natural product information in the KG, including additional natural products of interest and other knowledge domains. The NP-KG framework is publicly available at https://doi.org/10.5281/zenodo.6814507. The code for relation extraction, KG construction, and hypothesis generation is available at https://github.com/sanyabt/np-kg.

**Abbreviations**

| | |
|---|---|
| ChEBI | Chemical Entities of Biological Interest |
| CYP | Cytochrome P450 |
| DIKB | Drug Interaction Knowledge Base |
| ECG | (-)-epicatechin-3 gallate |
| EGCG | (-)-epigallocatechin gallate |
| FDA | Food and Drug Administration |
| INDRA | Integrated Network and Dynamical Reasoning Assembler |
| IVIVE | *In vitro* to *in vivo* extrapolation |
| KG | Knowledge Graph |
| NPDI | Natural Product-Drug Interaction |
| OATP | Human Organic Anion Transporting Polypeptide |
| OBO | Open Biological and Biomedical Ontology |
| OWL | Web Ontology Language |
| OWLNETS | Network Transformation for Statistical Learning |
| PheKnowLator | Phenotype Knowledge Translator |
| REACH | Reading and Assembling Contextual and Holistic Mechanisms from Text |
| UGT | uridine 5'-diphospho-glucuronosyltransferase (UDP-glucuronosyltransferase) |
| UMLS | Unified Medical Language System |
| US | United States |

**Funding**

This study was supported by the National Institutes of Health National Center for Complementary and Integrative Health and Office of Dietary Supplements (Grant U54 AT008909).

**Acknowledgments**

All figures were created with BioRender.com.

## Author Contributions

**Sanya B. Taneja:** Conceptualization, Methodology, Software, Validation, Visualization, Writing – Original Draft, Writing – Review and Editing; **Tiffany J. Callahan:** Methodology, Software, Writing – Review and Editing; **Mary F. Paine:** Conceptualization, Supervision, Writing – Review and Editing; **Sandra L. Kane-Gill:** Conceptualization, Writing – Review and Editing; **Halil Kilicoglu:** Writing – Review and Editing; **Marcin P. Joachimiak:** Writing – Review and Editing; **Richard D. Boyce:** Conceptualization, Methodology, Validation, Supervision, Writing – Review and Editing.

## References

[1] T.C. Clarke, L.I. Black, B.J. Stussman, P.M. Barnes, R.L. Nahin, Trends in the Use of Complementary Health Approaches Among Adults: United States, 2002–2012, Natl Health Stat Report. (2015) 1–16.

[2] T. Smith, F. Majid, V. Eckl, C.M. Reynolds, Herbal Supplement Sales in US Increase by Record-Breaking 17.3% in 2020, (n.d.) 14.

[3] B. Gurley, Pharmacokinetic Herb-Drug Interactions (Part 1): Origins, Mechanisms, and the Impact of Botanical Dietary Supplements, Planta Med. 78 (2012) 1478–1489. https://doi.org/10.1055/s-0031-1298273.

[4] S.J. Brantley, A.A. Argikar, Y.S. Lin, S. Nagar, M.F. Paine, Herb–Drug Interactions: Challenges and Opportunities for Improved Predictions, Drug Metab Dispos. 42 (2014) 301–317. https://doi.org/10.1124/dmd.113.055236.

[5] T.B. Agbabiaka, B. Wider, L.K. Watson, C. Goodman, Concurrent Use of Prescription Drugs and Herbal Medicinal Products in Older Adults: A Systematic Review, Drugs Aging. 34 (2017) 891–905. https://doi.org/10.1007/s40266-017-0501-7.

[6] C. Birer-Williams, B.T. Gufford, E. Chou, M. Alilio, S. VanAlstine, R.E. Morley, J.S. McCune, M.F. Paine, R.D. Boyce, A New Data Repository for Pharmacokinetic Natural Product-Drug Interactions: From Chemical Characterization to Clinical Studies, Drug Metab Dispos. 48 (2020) 1104–1112. https://doi.org/10.1124/dmd.120.000054.

[7] M.F. Paine, D.D. Shen, J.S. McCune, Recommended Approaches for Pharmacokinetic Natural Product-Drug Interaction Research: a NaPDI Center Commentary, Drug Metabolism and Disposition. 46 (2018) 1041. https://doi.org/10.1124/dmd.117.079962.

[8] S. Misaka, J. Yatabe, F. Müller, K. Takano, K. Kawabe, H. Glaeser, M.S. Yatabe, S. Onoue, J.P. Werba, H. Watanabe, S. Yamada, M.F. Fromm, J. Kimura, Green tea ingestion greatly reduces plasma concentrations of nadolol in healthy subjects, Clin Pharmacol Ther. 95 (2014) 432–438. https://doi.org/10.1038/clpt.2013.241.

[9] I.J. Tripodi, T.J. Callahan, J.T. Westfall, N.S. Meitzer, R.D. Dowell, L.E. Hunter, Applying knowledge-driven mechanistic inference to toxicogenomics, Toxicology in Vitro. 66 (2020) 104877. https://doi.org/10.1016/j.tiv.2020.104877.

[10] L. Wang, O. Tafjord, A. Cohan, S. Jain, S. Skjonsberg, C. Schoenick, N. Botner, W. Ammar, SUPP.AI: finding evidence for supplement-drug interactions, in: Proceedings of the 58th Annual Meeting of the Association for Computational Linguistics: System Demonstrations, Association for Computational Linguistics, Online, 2020: pp. 362–371. https://doi.org/10.18653/v1/2020.acl-demos.41.

[11] D. Schutte, J. Vasilakes, A. Bompelli, Y. Zhou, M. Fiszman, H. Xu, H. Kilicoglu, J.R. Bishop, T. Adam, R. Zhang, Discovering novel drug-supplement interactions using SuppKG generated from the biomedical literature, Journal of Biomedical Informatics. 131 (2022) 104120. https://doi.org/10.1016/j.jbi.2022.104120.

[12] M.M. Rahman, S.M. Vadrev, A. Magana-Mora, J. Levman, O. Soufan, A novel graph mining approach to predict and evaluate food-drug interactions, Sci Rep. 12 (2022) 1061. https://doi.org/10.1038/s41598-022-05132-y.

[13] M.F. Paine, Natural Products: Experimental Approaches to Elucidate Disposition Mechanisms and Predict Pharmacokinetic Drug Interactions, Drug Metab Dispos. 48 (2020) 956–962. https://doi.org/10.1124/dmd.120.000182.

[14] C. for D.E. and Research, Drug Interactions | Relevant Regulatory Guidance and Policy Documents, FDA. (2021). https://www.fda.gov/drugs/drug-interactions-labeling/drug-interactions-relevant-regulatory-guidance-and-policy-documents (accessed June 22, 2022).

[15] D.-D. Tian, J.J. Kellogg, N. Okut, N.H. Oberlies, N.B. Cech, D.D. Shen, J.S. McCune, M.F. Paine, Identification of Intestinal UDP-Glucuronosyltransferase Inhibitors in Green Tea (Camellia sinensis) Using a Biochemometric Approach: Application to Raloxifene as a Test Drug via In Vitro to In Vivo Extrapolation, Drug Metab Dispos. 46 (2018) 552–560. https://doi.org/10.1124/dmd.117.079491.

[16] J. McCune, D. Tian, P. Hardy, N. Cech, D. Shen, M. Layton, J. White, M. Paine, GREEN TEA DECREASES RALOXIFENE SYSTEMIC EXPOSURE TO BELOW THE PRE-DEFINED NO EFFECT RANGE IN HEALTHY VOLUNTEERS., in: CLINICAL PHARMACOLOGY & THERAPEUTICS, WILEY 111 RIVER ST, HOBOKEN 07030-5774, NJ USA, 2018: pp. S35–S35.

[17] DEA Announces Intent To Schedule Kratom, (n.d.). https://www.dea.gov/press-releases/2016/08/30/dea-announces-intent-schedule-kratom (accessed August 17, 2022).

[18] M. Anwar, R. Law, J. Schier, Notes from the Field: Kratom (Mitragyna speciosa) Exposures Reported to Poison Centers - United States, 2010-2015, MMWR Morb Mortal Wkly Rep. 65 (2016) 748–749. https://doi.org/10.15585/mmwr.mm6529a4.

[19] V.R. Ballotin, L.G. Bigarella, A.B. de M. Brandão, R.A. Balbinot, S.S. Balbinot, J. Soldera, Herb-induced liver injury: Systematic review and meta-analysis, World J Clin Cases. 9 (2021) 5490–5513. https://doi.org/10.12998/wjcc.v9.i20.5490.

[20] E.O. Olsen, J. O'Donnell, C.L. Mattson, J.G. Schier, N. Wilson, Notes from the field: unintentional drug overdose deaths with kratom detected—27 states, July 2016–December 2017, Morbidity and Mortality Weekly Report. 68 (2019) 326.

[21] R. Hoehndorf, P.N. Schofield, G.V. Gkoutos, The role of ontologies in biological and biomedical research: a functional perspective, Briefings in Bioinformatics. 16 (2015) 1069–1080. https://doi.org/10.1093/bib/bbv011.

[22] J. Hastings, G. Owen, A. Dekker, M. Ennis, N. Kale, V. Muthukrishnan, S. Turner, N. Swainston, P. Mendes, C. Steinbeck, ChEBI in 2016: Improved services and an expanding collection of metabolites, Nucleic Acids Res. 44 (2016) D1214-1219. https://doi.org/10.1093/nar/gkv1031.

[23] R. Jackson, N. Matentzoglu, J.A. Overton, R. Vita, J.P. Balhoff, P.L. Buttigieg, S. Carbon, M. Courtot, A.D. Diehl, D.M. Dooley, W.D. Duncan, N.L. Harris, M.A. Haendel, S.E. Lewis, D.A. Natale, D. Osumi-Sutherland, A. Ruttenberg, L.M. Schriml, B. Smith, C.J. Stoeckert Jr., N.A. Vasilevsky, R.L. Walls, J. Zheng, C.J. Mungall, B. Peters, OBO Foundry in 2021: operationalizing open data principles to evaluate ontologies, Database. 2021 (2021) baab069. https://doi.org/10.1093/database/baab069.

[24] T.J. Callahan, I.J. Tripodi, L.E. Hunter, W.A. Baumgartner, A Framework for Automated Construction of Heterogeneous Large-Scale Biomedical Knowledge Graphs, BioRxiv. (2020) 2020.04.30.071407. https://doi.org/10.1101/2020.04.30.071407.

[25] C.J. Mungall, J.A. McMurry, S. Köhler, J.P. Balhoff, C. Borromeo, M. Brush, S. Carbon, T. Conlin, N. Dunn, M. Engelstad, E. Foster, J.P. Gourdine, J.O.B. Jacobsen, D. Keith, B. Laraway, S.E. Lewis, J. NguyenXuan, K. Shefchek, N. Vasilevsky, Z. Yuan, N. Washington, H. Hochheiser, T. Groza, D. Smedley, P.N. Robinson, M.A. Haendel, The Monarch Initiative: an integrative data and analytic platform connecting phenotypes to genotypes across species, Nucleic Acids Res. 45 (2017) D712–D722. https://doi.org/10.1093/nar/gkw1128.

[26] S. Köhler, L. Carmody, N. Vasilevsky, J.O.B. Jacobsen, D. Danis, J.-P. Gourdine, M. Gargano, N.L. Harris, N. Matentzoglu, J.A. McMurry, D. Osumi-Sutherland, V. Cipriani, J.P. Balhoff, T. Conlin, H. Blau, G. Baynam, R. Palmer, D. Gratian, H. Dawkins, M. Segal, A.C. Jansen, A. Muaz, W.H. Chang, J. Bergerson, S.J.F. Lauledorkind, Z. Yüksel, S. Beltran, A.F. Freeman, P.I. Sergouniotis, D. Durkin, A.L. Storm, M. Hanauer, M. Brudno, S.M. Bello, M. Sincan, K. Rageth, M.T. Wheeler, R. Oegema, H. Lourghi, M.G. Della Rocca, R. Thompson, F. Castellanos, J. Priest, C. Cunningham-Rundles, A. Hegde, R.C. Lovering, C. Hajek, A. Olry, L. Notarangelo, M. Similuk, X.A. Zhang, D. Gómez-Andrés, H. Lochmüller, H. Dollfus, S. Rosenzweig, S. Marwaha, A. Rath, K. Sullivan, C. Smith, J.D. Milner, D. Leroux, C.F.

Boerkoel, A. Klion, M.C. Carter, T. Groza, D. Smedley, M.A. Haendel, C. Mungall, P.N. Robinson, Expansion of the Human Phenotype Ontology (HPO) knowledge base and resources, Nucleic Acids Res. 47 (2019) D1018–D1027. https://doi.org/10.1093/nar/gky1105.

[27] C.J. Mungall, C. Torniai, G.V. Gkoutos, S.E. Lewis, M.A. Haendel, Uberon, an integrative multi-species anatomy ontology, Genome Biology. 13 (2012) R5. https://doi.org/10.1186/gb-2012-13-1-r5.

[28] M. Ashburner, C.A. Ball, J.A. Blake, D. Botstein, H. Butler, J.M. Cherry, A.P. Davis, K. Dolinski, S.S. Dwight, J.T. Eppig, M.A. Harris, D.P. Hill, L. Issel-Tarver, A. Kasarskis, S. Lewis, J.C. Matese, J.E. Richardson, M. Ringwald, G.M. Rubin, G. Sherlock, Gene Ontology: tool for the unification of biology, Nat Genet. 25 (2000) 25–29. https://doi.org/10.1038/75556.

[29] D.A. Natale, C.N. Arighi, W.C. Barker, J.A. Blake, C.J. Bult, M. Caudy, H.J. Drabkin, P. D'Eustachio, A.V. Evsikov, H. Huang, J. Nchoutmboube, N.V. Roberts, B. Smith, J. Zhang, C.H. Wu, The Protein Ontology: a structured representation of protein forms and complexes, Nucleic Acids Res. 39 (2011) D539-545. https://doi.org/10.1093/nar/gkq907.

[30] V. Petri, P. Jayaraman, M. Tutaj, G.T. Hayman, J.R. Smith, J. De Pons, S.J. Laulederkind, T.F. Lowry, R. Nigam, S.-J. Wang, M. Shimoyama, M.R. Dwinell, D.H. Munzenmaier, E.A. Worthey, H.J. Jacob, The pathway ontology – updates and applications, J Biomed Semantics. 5 (2014) 7. https://doi.org/10.1186/2041-1480-5-7.

[31] C.J. Mungall, C. Batchelor, K. Eilbeck, Evolution of the Sequence Ontology terms and relationships, Journal of Biomedical Informatics. 44 (2011) 87–93. https://doi.org/10.1016/j.jbi.2010.03.002.

[32] J. Bard, S.Y. Rhee, M. Ashburner, An ontology for cell types, Genome Biol. 6 (2005) R21. https://doi.org/10.1186/gb-2005-6-2-r21.

[33] S. Sarntivijai, Y. Lin, Z. Xiang, T.F. Meehan, A.D. Diehl, U.D. Vempati, S.C. Schürer, C. Pang, J. Malone, H. Parkinson, Y. Liu, T. Takatsuki, K. Saijo, H. Masuya, Y. Nakamura, M.H. Brush, M.A. Haendel, J. Zheng, C.J. Stoeckert, B. Peters, C.J. Mungall, T.E. Carey, D.J. States, B.D. Athey, Y. He, CLO: The cell line ontology, J Biomed Semantics. 5 (2014) 37. https://doi.org/10.1186/2041-1480-5-37.

[34] [Dataset] T. Callahan, PheKnowLator, (2019). https://doi.org/10.5281/zenodo.3401437.

[35] Y. He, S. Sarntivijai, Y. Lin, Z. Xiang, A. Guo, S. Zhang, D. Jagannathan, L. Toldo, C. Tao, B. Smith, OAE: The Ontology of Adverse Events, J Biomed Semantics. 5 (2014) 29. https://doi.org/10.1186/2041-1480-5-29.

[36] R. Boyce, C. Collins, J. Horn, I. Kalet, Computing with evidence: Part I: A drug-mechanism evidence taxonomy oriented toward confidence assignment, Journal of Biomedical Informatics. 42 (2009) 979–989. https://doi.org/10.1016/j.jbi.2009.05.001.

[37] R. Boyce, C. Collins, J. Horn, I. Kalet, Computing with evidence Part II: An evidential approach to predicting metabolic drug-drug interactions, J Biomed Inform. 42 (2009) 990–1003. https://doi.org/10.1016/j.jbi.2009.05.010.

[38] S. Avram, C.G. Bologa, J. Holmes, G. Bocci, T.B. Wilson, D.-T. Nguyen, R. Curpan, L. Halip, A. Bora, J.J. Yang, J. Knockel, S. Sirimulla, O. Ursu, T.I. Oprea, DrugCentral 2021 supports drug discovery and repositioning, Nucleic Acids Research. 49 (2021) D1160–D1169. https://doi.org/10.1093/nar/gkaa997.

[39] C. Yeung, K. Yoshida, M. Kusama, H. Zhang, I. Ragueneau-Majlessi, S. Argon, L. Li, P. Chang, C. Le, P. Zhao, L. Zhang, Y. Sugiyama, S.-M. Huang, Organ Impairment—Drug–Drug Interaction Database: A Tool for Evaluating the Impact of Renal or Hepatic Impairment and Pharmacologic Inhibition on the Systemic Exposure of Drugs, CPT Pharmacometrics Syst Pharmacol. 4 (2015) 489–494. https://doi.org/10.1002/psp4.55.

[40] T.J. Callahan, W.A. Baumgartner, Jr, M. Bada, A.L. Stefanski, I. Tripodi, E.K. White, L.E. Hunter, OWL-NETS: Transforming OWL Representations for Improved Network Inference, Pacific Symposium on Biocomputing. Pacific Symposium on Biocomputing. 23 (2018) 133.

[41] A.A. Hagberg, D.A. Schult, P.J. Swart, Exploring Network Structure, Dynamics, and Function using NetworkX, in: G. Varoquaux, T. Vaught, J. Millman (Eds.), Proceedings of the 7th Python in Science Conference, Pasadena, CA USA, 2008: pp. 11–15.

[42] J. Hanna, E. Joseph, M. Brochhausen, W.R. Hogan, Building a drug ontology based on RxNorm and other sources, J Biomed Semantics. 4 (2013) 44. https://doi.org/10.1186/2041-1480-4-44.

[43] D.M. Dooley, E.J. Griffiths, G.S. Gosal, P.L. Buttigieg, R. Hoehndorf, M.C. Lange, L.M. Schriml, F.S. Brinkman, W.W. Hsiao, FoodOn: a harmonized food ontology to increase global food traceability, quality control and data integration, Npj Science of Food. 2 (2018) 1–10.

[44] C.L. Schoch, S. Ciufo, M. Domrachev, C.L. Hotton, S. Kannan, R. Khovanskaya, D. Leipe, R. Mcveigh, K. O'Neill, B. Robbertse, S. Sharma, V. Soussov, J.P. Sullivan, L. Sun, S. Turner, I. Karsch-Mizrachi, NCBI Taxonomy: a comprehensive update on curation, resources and tools, Database (Oxford). 2020 (2020) baaa062. https://doi.org/10.1093/database/baaa062.

[45] S.B. Taneja, T.J. Callahan, M. Brochhausen, M.F. Paine, S.L. Kane-Gill, R.D. Boyce, Designing potential extensions from G-SRS to ChEBI to identify natural product-drug interactions, in: 2021. https://doi.org/10.5281/zenodo.5736386.

[46] T. Peryea, N. Southall, M. Miller, D. Katzel, N. Anderson, J. Neyra, S. Stemann, Đ.-T. Nguyễn, D. Amugoda, A. Newatia, R. Ghazzaoui, E. Johanson, H. Diederik, L. Callahan, F. Switzer, Global Substance Registration System: consistent scientific descriptions for substances related to health, Nucleic Acids Research. 49 (2021) D1179–D1185. https://doi.org/10.1093/nar/gkaa962.

[47] H. Kilicoglu, G. Rosemblat, M. Fiszman, D. Shin, Broad-coverage biomedical relation extraction with SemRep, BMC Bioinformatics. 21 (2020) 188. https://doi.org/10.1186/s12859-020-3517-7.

[48] B.M. Gyori, J.A. Bachman, K. Subramanian, J.L. Muhlich, L. Galescu, P.K. Sorger, From word models to executable models of signaling networks using automated assembly, Molecular Systems Biology. 13 (2017) 954. https://doi.org/10.15252/msb.20177651.

[49] M.A. Valenzuela-Escárcega, Ö. Babur, G. Hahn-Powell, D. Bell, T. Hicks, E. Noriega-Atala, X. Wang, M. Surdeanu, E. Demir, C.T. Morrison, Large-scale automated machine reading discovers new cancer-driving mechanisms, Database (Oxford). 2018 (2018). https://doi.org/10.1093/database/bay098.

[50] "MetaMap Team," MetaMap - A Tool For Recognizing UMLS Concepts in Text, (2015). http://metamap.nlm.nih.gov/ (accessed June 4, 2015).

[51] Y. Shinyama, PDFMiner, (2022). https://github.com/euske/pdfminer (accessed April 22, 2022).

[52] L. Hoang, R.D. Boyce, N. Bosch, B. Stottlemyer, M. Brochhausen, J. Schneider, Automatically classifying the evidence type of drug-drug interaction research papers as a step toward computer supported evidence curation, AMIA Annual Symposium Proceedings. 2020 (2020) 554.

[53] ontoRunNER, (2021). https://github.com/monarch-initiative/ontorunner (accessed April 21, 2022).

[54] R. Zhang, D. Hristovski, D. Schutte, A. Kastrin, M. Fiszman, H. Kilicoglu, Drug repurposing for COVID-19 via knowledge graph completion, Journal of Biomedical Informatics. 115 (2021) 103696. https://doi.org/10.1016/j.jbi.2021.103696.

[55] M. Cafasso, CLIPS Python bindings, (2022). https://github.com/noxdafox/clipspy (accessed April 23, 2022).

[56] D.S. Himmelstein, A. Lizee, C. Hessler, L. Brueggeman, S.L. Chen, D. Hadley, A. Green, P. Khankhanian, S.E. Baranzini, Systematic integration of biomedical knowledge prioritizes drugs for repurposing, ELife. 6 (2017) e26726. https://doi.org/10.7554/eLife.26726.

[57] S. Misaka, O. Abe, T. Ono, Y. Ono, H. Ogata, I. Miura, Y. Shikama, M.F. Fromm, H. Yabe, K. Shimomura, Effects of single green tea ingestion on pharmacokinetics of nadolol in healthy volunteers, Br J Clin Pharmacol. 86 (2020) 2314–2318. https://doi.org/10.1111/bcp.14315.

[58] R.S. Tanna, D.-D. Tian, N.B. Cech, N.H. Oberlies, A.E. Rettie, K.E. Thummel, M.F. Paine, Refined Prediction of Pharmacokinetic Kratom-Drug Interactions: Time-Dependent Inhibition Considerations, J Pharmacol Exp Ther. 376 (2021) 64–73. https://doi.org/10.1124/jpet.120.000270.

[59] S.H. Kamble, A. Sharma, T.I. King, E.C. Berthold, F. León, P.K.L. Meyer, S.R.R. Kanumuri, L.R. McMahon, C.R. McCurdy, B.A. Avery, Exploration of cytochrome P450 inhibition mediated drug-drug interaction potential of kratom alkaloids, Toxicol Lett. 319 (2020) 148–154. https://doi.org/10.1016/j.toxlet.2019.11.005.

[60] H.D. Brogdon, M.M. McPhee, M.F. Paine, E.J. Cox, A.G. Burns, A Case of Potential Pharmacokinetic Kratom-drug Interactions Resulting in Toxicity and Subsequent Treatment of

Kratom Use Disorder With Buprenorphine/Naloxone, J Addict Med. (2022). https://doi.org/10.1097/ADM.0000000000000968.

[61] J. Vasilakes, R. Rizvi, G.B. Melton, S. Pakhomov, R. Zhang, Evaluating active learning methods for annotating semantic predications, JAMIA Open. 1 (2018) 275–282. https://doi.org/10.1093/jamiaopen/ooy021.

[62] H. Kilicoglu, D. Shin, M. Fiszman, G. Rosemblat, T.C. Rindflesch, SemMedDB: a PubMed-scale repository of biomedical semantic predications, Bioinformatics. 28 (2012) 3158–3160. https://doi.org/10.1093/bioinformatics/bts591.

[63] G. Rosemblat, M. Fiszman, D. Shin, H. Kilicoglu, Towards a characterization of apparent contradictions in the biomedical literature using context analysis, Journal of Biomedical Informatics. 98 (2019) 103275. https://doi.org/10.1016/j.jbi.2019.103275.

[64] D.N. Sosa, R.B. Altman, Contexts and contradictions: a roadmap for computational drug repurposing with knowledge inference, Briefings in Bioinformatics. (2022) bbac268. https://doi.org/10.1093/bib/bbac268.

[65] Q. Zhang, K. Yu, S. Li, X. Zhang, Q. Zhao, X. Zhao, Z. Liu, H. Cheng, Z.-X. Liu, X. Li, gutMEGA: a database of the human gut MEtaGenome Atlas, Brief Bioinform. 22 (2021) bbaa082. https://doi.org/10.1093/bib/bbaa082.

[66] X. Zeng, X. Yang, J. Fan, Y. Tan, L. Ju, W. Shen, Y. Wang, X. Wang, W. Chen, D. Ju, Y.Z. Chen, MASI: microbiota-active substance interactions database, Nucleic Acids Res. 49 (2021) D776–D782. https://doi.org/10.1093/nar/gkaa924.

[67] M.R. Meyer, L. Wagmann, N. Schneider-Daum, B. Loretz, C. de Souza Carvalho, C.-M. Lehr, H.H. Maurer, P-glycoprotein interactions of novel psychoactive substances - stimulation of ATP consumption and transport across Caco-2 monolayers, Biochem Pharmacol. 94 (2015) 220–226. https://doi.org/10.1016/j.bcp.2015.01.008.

[68] V.K. Manda, B. Avula, Z. Ali, I.A. Khan, L.A. Walker, S.I. Khan, Evaluation of in vitro absorption, distribution, metabolism, and excretion (ADME) properties of mitragynine, 7-hydroxymitragynine, and mitraphylline, Planta Med. 80 (2014) 568–576. https://doi.org/10.1055/s-0034-1368444.

[69] N. Rusli, A. Amanah, G. Kaur, M.I. Adenan, S.F. Sulaiman, H.A. Wahab, M.L. Tan, The inhibitory effects of mitragynine on P-glycoprotein in vitro, Naunyn Schmiedebergs Arch Pharmacol. 392 (2019) 481–496. https://doi.org/10.1007/s00210-018-01605-y.

[70] Y. Luo, X. Zhao, J. Zhou, J. Yang, Y. Zhang, W. Kuang, J. Peng, L. Chen, J. Zeng, A network integration approach for drug-target interaction prediction and computational drug repositioning from heterogeneous information, Nature Communications. 8 (2017) 573. https://doi.org/10.1038/s41467-017-00680-8.

[71] I. Tripodi, K.B. Cohen, L.E. Hunter, A semantic knowledge-base approach to drug-drug interaction discovery, in: 2017 IEEE International Conference on Bioinformatics and Biomedicine (BIBM), IEEE, Kansas City, MO, 2017: pp. 1123–1126. https://doi.org/10.1109/BIBM.2017.8217814.

[72] V. Sharma, L.F.F. Gelin, I.N. Sarkar, Identifying Herbal Adverse Events From Spontaneous Reporting Systems Using Taxonomic Name Resolution Approach, Bioinform Biol Insights. 14 (2020) 1177932220921350. https://doi.org/10.1177/1177932220921350.

[73] V. Sharma, I.N. Sarkar, Identifying natural health product and dietary supplement information within adverse event reporting systems, in: Biocomputing 2018, WORLD SCIENTIFIC, Kohala Coast, Hawaii, USA, 2018: pp. 268–279. https://doi.org/10.1142/9789813235533_0025.

[74] J.A. Vasilakes, R.F. Rizvi, J. Zhang, T.J. Adam, R. Zhang, Detecting Signals of Dietary Supplement Adverse Events from the CFSAN Adverse Event Reporting System (CAERS), AMIA Jt Summits Transl Sci Proc. 2019 (2019) 258–266.

[75] E.J. Cox, D.-D. Tian, J.D. Clarke, A.E. Rettie, J.D. Unadkat, K.E. Thummel, J.S. McCune, M.F. Paine, Modeling Pharmacokinetic Natural Product-Drug Interactions for Decision-Making: A NaPDI Center Recommended Approach, Pharmacol Rev. 73 (2021) 847–859. https://doi.org/10.1124/pharmrev.120.000106.

[76] E.J. Cox, A.E. Rettie, J.D. Unadkat, K.E. Thummel, J.S. McCune, M.F. Paine, Adapting regulatory drug-drug interaction guidance to design clinical pharmacokinetic natural product-

drug interaction studies: A NaPDI Center recommended approach, Clin Transl Sci. 15 (2022) 322–329. https://doi.org/10.1111/cts.13172.

[77] R.F. Rizvi, J. Vasilakes, T.J. Adam, G.B. Melton, J.R. Bishop, J. Bian, C. Tao, R. Zhang, iDISK: the integrated DIetary Supplements Knowledge base, J Am Med Inform Assoc. 27 (2020) 539–548. https://doi.org/10.1093/jamia/ocz216.

[78] S. Haussmann, O. Seneviratne, Y. Chen, Y. Ne'eman, J. Codella, C.-H. Chen, D.L. McGuinness, M.J. Zaki, FoodKG: A Semantics-Driven Knowledge Graph for Food Recommendation, in: C. Ghidini, O. Hartig, M. Maleshkova, V. Svátek, I. Cruz, A. Hogan, J. Song, M. Lefrançois, F. Gandon (Eds.), The Semantic Web – ISWC 2019, Springer International Publishing, Cham, 2019: pp. 146–162. https://doi.org/10.1007/978-3-030-30796-7_10.

[79] J.Y. Ryu, H.U. Kim, S.Y. Lee, Deep learning improves prediction of drug–drug and drug–food interactions, Proceedings of the National Academy of Sciences. 115 (2018) E4304–E4311. https://doi.org/10.1073/pnas.1803294115.

# Developing a Knowledge Graph Framework for Pharmacokinetic Natural Product-Drug Interactions

# Appendix A

### Search Strategies

Search strategies for green tea and kratom were adapted from the following:

Birer-Williams C, Gufford BT, Chou E, Alilio M, VanAlstine S, Morley RE, McCune JS, Paine MF, Boyce RD. A new data repository for pharmacokinetic natural product-drug interactions: from chemical characterization to clinical studies. Drug metabolism and disposition. 2020 Oct 1;48(10):1104-12.

### Search Strategies for Green tea

**Clinical Studies**
(Clinical Trial [PT] AND ("green tea"[All Fields] OR "green teas"[All Fields] OR "green tea extract"[All Fields] OR "Camellia sinensis"[All Fields] OR "Cha ye"[All Fields] OR "Cha-yeh"[All Fields] OR "Camelliae folium"[All Fields] OR "Camelliae sinensis folium "[All Fields] OR "Matcha tea"[All Fields] OR "Sencha tea"[All Fields] OR "Thea sinensis"[All Fields] OR "Catechin"[All Fields] OR "Catechins"[All Fields] OR "epicatechin"[All Fields] OR "gallocatechin"[All Fields] OR "epigallocatechin"[All Fields] OR "epigallocatechin gallate"[All Fields] OR "EGCG"[All Fields] OR "Epicatechin-3-gallate"[All Fields] OR "(-)-epigallocatechin gallate"[All Fields] OR "(-)-epicatechin"[All Fields] OR "(-)-epigallocatechin"[All Fields] OR "(-)-epicatechin-3-gallate"[All Fields]) AND "drug interactions"[All Fields]) NOT Review [PT]

**Case Reports**
Case Reports [PT] AND ("green tea"[All Fields] OR "green teas"[All Fields] OR "green tea extract"[All Fields] OR "Camellia sinensis"[All Fields] OR "Cha ye"[All Fields] OR "Cha-yeh"[All Fields] OR "Camelliae folium"[All Fields] OR "Camelliae sinensis folium "[All Fields] OR "Matcha tea"[All Fields] OR "Sencha tea"[All Fields] OR "Thea sinensis"[All Fields] OR "Catechin"[All Fields] OR "Catechins"[All Fields] OR "epicatechin"[All Fields] OR "gallocatechin"[All Fields] OR "epigallocatechin"[All Fields] OR "epigallocatechin gallate"[All Fields] OR "EGCG"[All Fields] OR "Epicatechin-3-gallate"[All Fields] OR "(-)-epigallocatechin gallate"[All Fields] OR "(-)-epicatechin"[All Fields] OR "(-)-epigallocatechin"[All Fields] OR "(-)-epicatechin-3-gallate"[All Fields]) AND "drug interactions"[All Fields]

**Mechanistic Studies**
Step 1) Log into My NCBI and go to Pubmed: https://www.ncbi.nlm. nih.gov/pubmed/.
Step 2) In the advanced search form, clear the search history.
Step 3) Paste in this query into builder (using "edit") and click "add to history"—this is referred to as "#1" in the rest of this search strategy:
"Cytochrome P-450 Enzyme System"[MeSH Terms] OR "Cyto- chrome P450 Family 1"[MeSH Terms] OR "Cytochrome P450 Family 2"[MeSH Terms] OR "Cytochrome P450 Family 3"[MeSH Terms] OR CYP1A1[All Fields] OR CYP1A2[All Fields] OR CYP1A3[All Fields] OR CYP1A4[All Fields] OR CYP1A5[All Fields] OR CYP2D6[All Fields] OR CYP2C9[All Fields] OR CYP2A6[All Fields] OR CYP2C8 [All Fields] OR CYP2C19[All Fields] OR CYP2B6[All Fields] OR CYP2B1[All Fields] OR CYP2E1[All Fields] OR CYP3A4[All Fields] OR CYP3A5[All Fields] OR UGT1[All Fields] OR UGT1A1[All Fields] OR UGT1A3[All Fields] OR UGT1A4[All Fields] OR UGT1A5[All Fields] OR UGT1A6[All Fields] OR UGT1A7[All Fields] OR UGT1A8[All Fields] OR UGT1A9[All Fields] OR UGT1A10[All Fields] OR UGT2[All Fields] OR UGT2A1[All Fields] OR UGT2A2 [All Fields] OR UGT2A3[All Fields] OR UGT2B4[All Fields] OR UGT2B7[All Fields] OR UGT2B10[All Fields] OR UGT2B11[All Fields] OR UGT2B15[All Fields] OR UGT2B17[All Fields] OR UGT2B28[All Fields] OR B3GAT1[All Fields] OR B3GAT2[All Fields] OR B3GAT3[All Fields].
Step 4) Paste in this query into builder (using "edit") and click "add to history"—this is referred to as "#2" in the rest of this search strategy.
"Solute Carrier Proteins"[MeSH Terms] OR "Membrane Transport Proteins"[MeSH Terms] OR "P-gp"[All Fields] OR "p-glycoprotei- n"[All Fields] OR BCRP[All Fields] OR OCT2[All Fields] OR

"Organic Cation Transporter 2"[MeSH Terms] OR "Organic Cation Transport Proteins"[MeSH Terms] OR MATE1[All Fields] OR "SLC4A Proteins"[MeSH Terms] OR MATE-2K[All Fields] OR "SLC4A Proteins"[MeSH Terms] OR OATP[All Fields] OR OAT1 [All Fields] OR "Organic Anion Transport Protein 1"[MeSH Terms] OR OAT3[All Fields] OR UGT1[All Fields] OR "Glucuronosyltransferase"[MeSH Terms] OR ABC[All Fields] OR "ATP-Binding Cassette Transporters"[MeSH Terms].
Step 5) Paste in this query into builder (using "edit") and click "add to history"—this is referred to as "#3" in the rest of this search strategy.

("green tea"[All Fields] OR "green teas"[All Fields] OR "green tea extract"[All Fields] OR "Camellia sinensis"[All Fields] OR "Cha ye"[All Fields] OR "Cha-yeh"[All Fields] OR "Camelliae folium"[All Fields] OR "Camelliae sinensis folium "[All Fields] OR "Matcha tea"[All Fields] OR "Sencha tea"[All Fields] OR "Thea sinensis"[All Fields] OR "Catechin"[All Fields] OR "Catechins"[All Fields] OR "epicatechin"[All Fields] OR "gallocatechin"[All Fields] OR "epigallocatechin"[All Fields] OR "epigallocatechin gallate"[All Fields] OR "EGCG"[All Fields] OR "Epicatechin-3-gallate"[All Fields] OR "(-)-epigallocatechin gallate"[All Fields] OR "(-)-epicatechin"[All Fields] OR "(-)-epigallocatechin"[All Fields] OR "(-)-epicatechin-3-gallate"[All Fields])
Step 6) Paste in this query into builder (using "edit") and click "add to history"—this is referred to as "#4" in the rest of this search strategy.

(Pharmacokinetics[MeSH Terms] OR pharmacokinetic[All Fields]) or (inhibit[All Fields] or inhibition[All Fields]) OR substrate[All Fields]. Step 7) Paste in this query into builder (using "edit") and click "add to history"—this is referred to as "#5" in the rest of this search strategy.

#3 AND #4 AND (#1 OR #2)

**Search Strategies for Kratom**

**Clinical Studies**
(Clinical Trial [PT] AND ("Kratom"[All Fields] OR "mitragynine"[All Fields] OR "mitragynine ethanedisulfonate"[All Fields] OR "SK and F 12711"[All Fields] OR "SKF 12711"[All Fields] OR "SK and F-12711"[All Fields] OR "mitragynine, (16E)-isomer"[All Fields] OR "mitragynine, (3beta,16E)-isomer"[All Fields] OR "mmitragynine, (3beta,16E,20beta)-isomer"[- All Fields] OR "kratom alkaloids"[All Fields] OR "kmitragynine monohydrochloride"[All Fields] OR "Mitragyna speciosa"[All Fields] OR "Nauclea speciose"[All Fields] OR "Biak-biak"[All Fields] OR "Cratom"[All Fields] OR "Gratom"[All Fields] OR "Ithang"[All Fields] OR "Kakuam"[All Fields] OR "Katawn"[All Fields] OR "Kedem- ba"[All Fields] OR "Ketum"[All Fields] OR "Krathom"[All Fields] OR "Kraton"[All Fields] OR "Kratum"[All Fields] OR "Madat"[All Fields] OR "Mambog"[All Fields] OR "Mitragynine"[All Fields] OR "Mitra- gynine extract"[All Fields] OR "Thang"[All Fields] OR "Thom"[All Fields] OR "7-hydroxymitragynine"[All Fields] OR "7-hydroxy-mitra- gynine"[All Fields] OR "mitragynine pseudoindoxyl"[All Fields] OR "Paynantheine"[All Fields]) AND "drug interactions"[All Fields]) NOT Review [PT]

**Case Reports**
Case Reports [PT] AND ("Kratom"[All Fields] OR "mitragynine"[All Fields] OR "mitragynine ethanedisulfonate"[All Fields] OR "SK and F 12711"[All Fields] OR "SKF 12711"[All Fields] OR "SK and F-12711"[All Fields] OR "mitragynine, (16E)-isomer"[All Fields] OR "mitragynine, (3beta,16E)-isomer"[All Fields] OR "mmitragynine, (3beta,16E,20beta)-isomer"[All Fields] OR "kratom alkaloids"[All Fields] OR "kmitragynine monohydrochloride"[All Fields] OR "Mitragyna speciosa"[All Fields] OR "Nauclea speciose"[All Fields] OR "Biak-biak"[All Fields] OR "Cratom"[All Fields] OR "Gratom"[All Fields] OR "Ithang"[All Fields] OR "Kakuam"[All Fields] OR "Katawn"[All Fields] OR "Kedemba"[All Fields] OR "Ketum"[All Fields] OR "Krathom"[All Fields] OR "Kraton"[All Fields] OR "Kratum"[All Fields] OR "Madat"[All Fields] OR "Mambog"[All Fields] OR "Mitragynine"[All Fields] OR "Mitragynine extract"[All Fields] OR "Thang"[All Fields] OR "Thom"[All Fields] OR "7-hydroxymitragynine"[All Fields] OR "7-hydroxy-mitragynine"[All Fields] OR "mitragynine pseudoindoxyl"[All Fields] OR "Paynantheine"[All Fields]) AND "drug interactions"[All Fields]

**Mechanistic Studies**
Step 1) Log into My NCBI and go to Pubmed: https://www.ncbi.nlm. nih.gov/pubmed/.
Step 2) In the advanced search form, clear the search history.
Step 3) Paste in this query into builder (using "edit") and click "add to history"—this is referred to as

"#1" in the rest of this search strategy:

"Cytochrome P-450 Enzyme System"[MeSH Terms] OR "Cyto- chrome P450 Family 1"[MeSH Terms] OR "Cytochrome P450 Family 2"[MeSH Terms] OR "Cytochrome P450 Family 3"[MeSH Terms] OR CYP1A1[All Fields] OR CYP1A2[All Fields] OR CYP1A3[All Fields] OR CYP1A4[All Fields] OR CYP1A5[All Fields] OR CYP2D6[All Fields] OR CYP2C9[All Fields] OR CYP2A6[All Fields] OR CYP2C8 [All Fields] OR CYP2C19[All Fields] OR CYP2B6[All Fields] OR CYP2B1[All Fields] OR CYP2E1[All Fields] OR CYP3A4[All Fields] OR CYP3A5[All Fields] OR UGT1[All Fields] OR UGT1A1[All Fields] OR UGT1A3[All Fields] OR UGT1A4[All Fields] OR UGT1A5[All Fields] OR UGT1A6[All Fields] OR UGT1A7[All Fields] OR UGT1A8[All Fields] OR UGT1A9[All Fields] OR UGT1A10[All Fields] OR UGT2[All Fields] OR UGT2A1[All Fields] OR UGT2A2 [All Fields] OR UGT2A3[All Fields] OR UGT2B4[All Fields] OR UGT2B7[All Fields] OR UGT2B10[All Fields] OR UGT2B11[All Fields] OR UGT2B15[All Fields] OR UGT2B17[All Fields] OR UGT2B28[All Fields] OR B3GAT1[All Fields] OR B3GAT2[All Fields] OR B3GAT3[All Fields].

Step 4) Paste in this query into builder (using "edit") and click "add to history"—this is referred to as "#2" in the rest of this search strategy.

"Solute Carrier Proteins"[MeSH Terms] OR "Membrane Transport Proteins"[MeSH Terms] OR "P-gp"[All Fields] OR "p-glycoprotei- n"[All Fields] OR BCRP[All Fields] OR OCT2[All Fields] OR "Organic Cation Transporter 2"[MeSH Terms] OR "Organic Cation Transport Proteins"[MeSH Terms] OR MATE1[All Fields] OR "SLC4A Proteins"[MeSH Terms] OR MATE-2K[All Fields] OR "SLC4A Proteins"[MeSH Terms] OR OATP[All Fields] OR OAT1 [All Fields] OR "Organic Anion Transport Protein 1"[MeSH Terms] OR OAT3[All Fields] OR UGT1[All Fields] OR "Glucuronosyltransfer- ase"[MeSH Terms] OR ABC[All Fields] OR "ATP-Binding Cassette Transporters"[MeSH Terms].

Step 5) Paste in this query into builder (using "edit") and click "add to history"—this is referred to as "#3" in the rest of this search strategy.

("Kratom"[All Fields] OR "mitragynine"[All Fields] OR "mitragynine ethanedisulfonate"[All Fields] OR "SK and F 12711"[All Fields] OR "SKF 12711"[All Fields] OR "SK and F-12711"[All Fields] OR "mitragynine, (16E)-isomer"[All Fields] OR "mitragynine, (3beta,16E)-isomer"[All Fields] OR "mmi- tragynine, (3beta,16E,20beta)-isomer"[All Fields] OR "kratom alka- loids"[All Fields] OR "kmitragynine monohydrochloride"[All Fields] OR "Mitragyna speciosa"[All Fields] OR "Nauclea speciose"[All Fields] OR "Biak-biak"[All Fields] OR "Cratom"[All Fields] OR "Gratom"[All Fields] OR "Ithang"[All Fields] OR "Kakuam"[All Fields] OR "Kataw- n"[All Fields] OR "Kedemba"[All Fields] OR "Ketum"[All Fields] OR "Krathom"[All Fields] OR "Kraton"[All Fields] OR "Kratum"[All Fields] OR "Madat"[All Fields] OR "Mambog"[All Fields] OR "Mitragynine"[All Fields] OR "Mitragynine extract"[All Fields] OR "Than- g"[All Fields] OR "Thom"[All Fields] OR "7-hydroxymitragynine"[All Fields] OR "7-hydroxy-mitragynine"[All Fields] OR "mitragynine pseudoindoxyl"[All Fields] OR "Paynantheine"[All Fields]).

Step 6) Paste in this query into builder (using "edit") and click "add to history"—this is referred to as "#4" in the rest of this search strategy.

(Pharmacokinetics[MeSH Terms] OR pharmacokinetic[All Fields]) or (inhibit[All Fields] or inhibition[All Fields]) OR substrate[All Fields]. Step 7) Paste in this query into builder (using "edit") and click "add to history"—this is referred to as "#5" in the rest of this search strategy.

#3 AND #4 AND (#1 OR #2)

**Table A.1. Edges in the ontology-grounded KG with edge counts, edge labels, and edge data source(s).**

| Edge Type | Edge Count | Edge Label | Edge Source(s) |
|---|---|---|---|
| **chemical-disease** | 171,506 | substance that treats | BioPortal, CTD |
| **chemical-gene** | 16,701 | interacts with | BioPortal, CTD |
| **chemical-biological process** | 304,686 | molecularly interacts with | BioPortal, CTD |
| **chemical-molecular** | 26,788 | molecularly interacts with | BioPortal, CTD |

| | function | | | |
|---|---|---|---|---|
| | chemical-cellular component | 46,372 | molecularly interacts with | BioPortal, CTD |
| | chemical-pathway | 29,248 | participates in | Reactome Pathway Database |
| | chemical-protein | 66,828 | interacts with | BioPortal, CTD |
| | chemical-transporter | 90 | transports | FDA Drug Interaction database |
| | chemical-molecule | 391 | molecularly interacts with | FDA Drug Interaction database, DIKB, Drug Central |
| | chemical-inhibitor | 272 | inhibits | FDA Drug Interaction database, DIKB |
| | chemical-substrate | 514 | substrate of | FDA Drug Interaction database, DIKB, Drug Central |
| | gene-pathway | 107,025 | participates in | CTD |
| | gene-phenotype | 23,525 | causes or contributes to | DisGeNET |
| | gene-protein | 19,523 | has gene product | UniProt |
| | pathway-cellular component | 15,982 | has component | Reactome Pathway Database |
| | pathway-molecular function | 2,424 | has function | Reactome Pathway Database |
| | protein-anatomy | 30,682 | located in | GTEx, Human Protein Atlas |
| | protein-cell | 75,318 | located in | GTEx, Human Protein Atlas |
| | protein-biological process | 129,408 | participates in | UniProt |
| | protein-molecular function | 70,085 | has function | UniProt |
| | protein-cellular component | 82,366 | located in | UniProt |
| | protein-pathway | 118,158 | participates in | Reactome Pathway Database |
| | protein-protein | 618,069 | molecularly interacts with | STRING database |

**Table A.2. Edges in the literature-based graph with edge counts, edge source(s), OBO label, and OBO identifier (N=25,421).**

| Edge Name | Edge Count (%) | Edge Source(s) | OBO Label | OBO Identifier |
|---|---|---|---|---|
| acetylation | 3 (0.012) | INDRA/REACH | protein acetylation | GO_0006473 |
| activation | 2515 (9.893) | INDRA/REACH | directly regulates activity of | RO_0002448 |
| affects | 353 (1.389) | SemRep | capable of regulating | RO_0002596 |
| associated_with | 110 (0.433) | SemRep | correlated_with | RO_0002610 |
| augments | 236 (0.928) | SemRep | capable of positively regulating | RO_0002598 |
| causes | 126 (0.496) | SemRep | causally influences | RO_0002566 |
| coexists_with | 1434 (5.641) | SemRep | existence overlaps | RO_0002490 |
| complicates | 1 (0.004) | SemRep | exacerbates condition | RO_0003309 |
| decrease_amount | 354 (1.393) | INDRA/REACH | directly negatively regulates quantity of | RO_0011010 |
| dehydroxylation | 5 (0.02) | INDRA/REACH | molecularly interacts with | RO_0002436 |
| demethylation | 3 (0.012) | INDRA/REACH | protein demethylation | GO_0006482 |
| dephosphorylation | 25 (0.098) | INDRA/REACH | protein dephosphorylation | GO_0006470 |
| deubiquitination | 1 (0.004) | INDRA/REACH | protein deubiquitination | GO_0016579 |
| disrupts | 153 (0.602) | SemRep | negatively regulates | RO_0002212 |
| glycosylation | 6 (0.024) | INDRA/REACH | protein glycosylation | GO_0006486 |
| hydroxylation | 8 (0.031) | INDRA/REACH | protein hydroxylation | GO_0018126 |
| increase_amount | 437 (1.719) | INDRA/REACH | directly positively regulates quantity of | RO_0011009 |
| inhibits | 3326 (13.084) | SemRep, INDRA/REACH | inhibits | RO_0002449 |
| interacts_with | 2421 (9.524) | SemRep | interacts with | RO_0002434 |
| methylation | 7 (0.028) | INDRA/REACH | protein methylation | GO_0006479 |
| part_of | 1927 (7.58) | SemRep | part of | BFO_0000050 |
| phosphorylation | 64 (0.252) | INDRA/REACH | phosphorylates | RO_0002447 |
| precedes | 1 (0.004) | SemRep | precedes | BFO_0000063 |
| predisposes | 27 (0.106) | SemRep | causes or contributes to condition | RO_0003302 |
| prevents | 53 (0.208) | SemRep | capable of inhibiting or preventing pathological process | RO_0002599 |
| produces | 248 (0.976) | SemRep | produces | RO_0003000 |
| **stimulates** | 11071 (43.551) | SemRep | positively regulates | RO_0002213 |
| **treats** | 505 (1.987) | SemRep | is substance that treats | RO_0002606 |
| **ubiquitination** | 1 (0.004) | INDRA/REACH | ubiquitinates | RO_0002480 |

**Table A.3.** Results from meta-path discovery in NP-KG for natural product-drug pairs. Meta-path descriptions are available in Figure 4.

| Meta-path | Results |
|---|---|
| **Green Tea - Raloxifene** | |
| **Green tea \| Green tea constituent** - *Relation* - **Raloxifene** | 1. EGCG inhibits raloxifene<br>2. EGCG coexists with raloxifene<br>3. EGCG interacts with raloxifene |
| **Raloxifene** - *Relation* - **Green tea \| Green tea constituent** | No results |
| **Green tea \| Green tea constituent** - *Relation A* - **Enzyme or Transporter** - *Relation B* - **Raloxifene** | No results |
| **Green tea \| Green tea constituent** - *Relation A* - **Enzyme or Transporter**, **Raloxifene** - *Relation B* - **Enzyme or Transporter** | 1. Green tea inhibits UGT, Raloxifene inhibits/interacts with UGT<br>2. ECG inhibits UGT, Raloxifene inhibits/interacts with UGT<br>3. Catechin positively regulates/interacts with/directly regulates activity of UGT, Raloxifene inhibits/interacts with UGT<br>4. Catechin positively regulates CYP3A4, Raloxifene inhibits CYP3A4<br>5. EGCG inhibits/positively regulates UGT, Raloxifene inhibits/interacts with UGT<br>6. EGCG positively regulates CYP3A4, Raloxifene inhibits/interacts with UGT |
| **Green Tea - Nadolol** | |
| **Green tea \| Green tea constituent** - Relation - **Nadolol** | 1. Green tea inhibits nadolol<br>2. EGCG inhibits nadolol<br>3. EGCG directly regulates activity of nadolol<br>4. Catechin interacts with nadolol |
| **Nadolol** - *Relation* – **Green tea \| Green tea constituent** | 1. Nadolol positively regulates EGCG<br>2. Nadolol interacts with EGCG<br>3. Nadolol positively regulates catechin<br>4. Nadolol positively regulates green tea leaf |
| **Green tea \| Green tea constituent** - *Relation A* – **Enzyme or Transporter** - *Relation B* – **Nadolol** | SLCO1A2 gene (OATP1A2) |
| **Green tea \| Green tea constituent** - *Relation A* – **Enzyme or Transporter**, **Nadolol** - *Relation B* – **Enzyme or Transporter** | See Tables A.4 and A.5. |
| **Kratom - Midazolam** | |
| **Kratom \| Kratom constituent** – Relation - **Midazolam** | No results |
| **Midazolam** - *Relation* – **Kratom \| Kratom constituent** | 1. Mitragynine inhibits midazolam<br>2. Mitragynine positively regulates midazolam |
| **Kratom \| Kratom constituent** – *Relation A* – **Enzyme or Transporter** - *Relation B* – **Midazolam** | No results |
| **Kratom \| Kratom constituent** – *Relation A* – **Enzyme or Transporter**, **Midazolam** - *Relation B* – **Enzyme or Transporter** | 1. Mitragynine interacts with/inhibits/directly regulates activity of/directly positively regulates quantity of CYP3A4, Midazolam is substrate of/interacts with CYP3A4<br>2. Mitragynine inhibits/interacts with/positively regulates CYP2D6, Midazolam inhibits CYP2D6<br>3. Mitragynine inhibits/directly |
| | negatively regulates quantity of ABCB1, Midazolam interacts with ABCB1 |
| **Kratom - Quetiapine** | |
| **Kratom | Kratom constituent** – Relation – **Quetiapine** | No results |
| **Quetiapine** – *Relation* – **Kratom | Kratom constituent** | No results |
| **Kratom | Kratom constituent** – *Relation A* – **Enzyme or Transporter** – *Relation B* – **Quetiapine** | No results |
| **Kratom | Kratom constituent** – *Relation A* – **Enzyme or Transporter**, **Quetiapine** – *Relation B* – **Enzyme or Transporter** | 1. Mitragynine interacts with/inhibits/directly regulates activity of/directly positively regulates quantity of CYP3A4, Quetiapine is substrate of CYP3A4<br>2. Mitragynine inhibits/directly negatively regulates quantity of P-glycoprotein, Quetiapine transports P-glycoprotein |
| **Kratom - Venlafaxine** | |
| **Kratom | Kratom constituent** – Relation – **Venlafaxine** | No results |
| **Venlafaxine** – *Relation* – **Kratom | Kratom constituent** | No results |
| **Kratom | Kratom constituent** – *Relation A* – **Enzyme or Transporter** – *Relation B* – **Venlafaxine** | No results |

| Kratom \| Kratom constituent – *Relation A* – **Enzyme or Transporter**, **Venlafaxine** - *Relation B* – **Enzyme or Transporter** | 1. Mitragynine interacts with/inhibits/directly regulates activity of/directly positively regulates quantity of CYP3A4, Venlafaxine is substrate of CYP3A4<br>2. Mitragynine inhibits/interacts with/positively regulates CYP2D6, Venlafaxine is substrate of/inhibits CYP2D6 |

**Table A.4.** Enzymes identified from meta-path discovery in NP-KG for green tea-nadolol.

| Enzyme | OBO Identifier |
|---|---|
| cytochrome P450 1A1 (CYP1A1) | PR_000006101 |
| cytochrome P450 1A2 (CYP1A2) | PR_000006102 |
| cytochrome P450 2D6 (CYP2D6) | PR_000006121 |
| cytochrome P450 2E1 (CYP2E1) | PR_000006122 |
| cytochrome P450 3A4 (CYP3A4) | PR_000006130 |
| cytochrome P450 7A1 (CYP7A1) | PR_000006148 |
| sulfotransferase 1A1 (SULT1A1) | PR_000015818 |
| UDP-glucuronosyltransferase 1A1 (UGT1A1) | PR_000017048 |

**Table A.5.** Transporters identified from meta-path discovery in NP-KG for green tea-nadolol.

| Transporter | OBO Identifier |
|---|---|
| high affinity copper uptake protein 1 (SLC31A1) | PR_000015083 |
| ileal sodium/bile acid cotransporter (IBAT / SLC10A2) | PR_000014918 |
| long-chain fatty acid transport protein 4 (FATP-4 / SLC27A4) | |
| multidrug and toxin extrusion protein 2 (SLC47A2 / MATE2) | PR_000015153 |
| multidrug resistance-associated protein 4 (MOAT-B / ABCC4) | PR_000003560 |
| phospholipid-transporting ATPase ABCA1 | PR_000003537 |
| phospholipid-transporting ATPase IB | PR_000029291 |
| (ATP8A2) | |
| sodium-dependent dopamine transporter (DA transporter / SLC6A3) | PR_Q01959 |
| solute carrier family 22 member 6 (SLC22A6 / OAT1) | PR_000014993 |
| solute carrier organic anion transporter family member 1A2 (SLCO1A2 / OATP1A2) | PR_000015222 |
| solute carrier organic anion transporter family member 1B3 (SLCO1B3 / OATP1B3) | PR_000015224 |
| zinc transporter ZIP4 (SLC39A4) | PR_000015131 |